# Using Curiosity for an Even Representation of Tasks in Continual Offline Reinforcement Learning

Pankayaraj Pathmanathan[a,*], Natalia Díaz-Rodríguez[b,c], Javier Del Ser[d,e]

[a]*Department of Computer Science, University of Maryland College Park, 20740, USA*
[b]*Department of Computer Science and Artificial Intelligence, University of Granada, CITIC, 18071 Granada, Spain*
[c]*Granada AI Lab, DaSCI Andalusian Institute in Data Science and Computational Intelligence, University of Granada, 18071 Granada, Spain*
[d]*TECNALIA, Basque Research & Technology Alliance (BRTA), 48160 Derio, Spain*
[e]*University of the Basque Country (UPV/EHU), 48013 Bilbao, Spain*

**Abstract**

***INTRODUCTION.*** In this work, we investigate the means of using curiosity on replay buffers to improve *offline multi-task* continual reinforcement learning when tasks, which are defined by the non-stationarity in the environment, are *non labeled* and *not evenly exposed to the learner in time*.

***METHODS.*** In particular, we investigate the use of curiosity both as a *tool for task boundary detection* and as a *priority metric when it comes to retaining old transition tuples*, which we respectively use to propose two different buffers. Firstly, we propose a Hybrid Reservoir Buffer with Task Separation (HRBTS), where curiosity is used to detect task boundaries that are not known due to the task agnostic nature of the problem. Secondly, by using curiosity as a priority metric when it comes to retaining old transition tuples, a Hybrid Curious Buffer (HCB) is proposed. We ultimately show that these buffers, in conjunction with regular reinforcement learning algorithms, can be used to alleviate the catastrophic forgetting issue suffered by the state of the art on replay buffers when the agent's exposure to tasks is not equal along time.

***RESULTS.*** We evaluate catastrophic forgetting and the efficiency of our proposed buffers against the latest works such as the Hybrid Reservoir Buffer (HRB) and the Multi-Time Scale Replay Buffer (MTR) in three different continual reinforcement learning settings. These settings are defined based on how many times the agent encounters the same task, how long they last, and how different new tasks are when compared to the old ones (i.e., how large the *task drift* is). The three settings are namely, 1. *prolonged task encounter with substantial task drift, and no task re-visitation*, 2. *frequent, short-lived task encounter with substantial task drift and task re-visitation*, and 3. *every timestep task encounter with small task drift and task re-visitation*. Experiments were done on classical control tasks and Metaworld environment.

***CONCLUSION.*** Experiments show that our proposed replay buffers display better immunity to catastrophic forgetting compared to existing works in all but the every time step task encounter with small task drift and task re-visitation. In this scenario curiosity will always be higher, thus not being an useful measure in both proposed buffers, making them not universally better than other approaches across all types of CL settings, and thereby opening up an avenue for further research.

*Keywords:* Reinforcement Learning, Continual Learning, Curiosity, Off-line learning

# 1. Introduction

Continual learning (CL) consists of the process of incrementally learning multiple tasks while exploiting and remembering the previously acquired knowledge [1] [2] [3]. These tasks can vary from having mere distributional shifts to target different and unique goals. Reinforcement learning (RL) [4], on the other hand, is a paradigm that deals with the issue of learning to solve a Markov Decision Process (MDP) in the context of a stationary environment. In many ways, even basic reinforcement learning can be seen through a continual learning framework. For example, when encountered with complex learning environments agents will not have access to all data at once and in some cases can also experience partial observability in terms of state observation, thus resulting in a distributional shift in the agent's learning data. This in itself can be considered a continual learning problem. Furthermore, when the environment itself changes over time, exhibiting non-stationarity, learning to generalize across multiple tasks becomes even more challenging.

As the agent encounters new tasks in sequence, the algorithm loses knowledge about the previously encountered and learned tasks, thus resulting in the agent failing to generalize for the past tasks while it tries to cope with the present. This phenomenon is commonly referred to as *catastrophic forgetting*, and it makes the agent necessitate to either save some actual data so that it can re-train on it along with the newly acquired samples, or preserve the old knowledge learned by the network by being conservative about the updates made to the model, as it encounters new tasks.

In the context of continual reinforcement learning, tasks can be characterized into two categories based on how they are encountered [5]. On one hand, new tasks can originate as a result of the agent's limited knowledge about the environment. While the environment remains stationary, due to the agent's lack of knowledge the agent can perceive a new region on the environment as a new task. Alternatively, these tasks can also be a result of the evolution or change in the environment's stationarity. In the case of the latter, these tasks can happen either with or without explicit labeling. For further reading about this categorization, we refer the reader to the survey in [5].

In cases where there is no explicit labeling (task agnostic RL), usual methods that capture knowledge consolidated thanks to the signaling of task boundaries [6, 7] become non-viable. In addition, these tasks can persist for different periods of time, resulting in a temporal discrepancy between tasks.

Intrinsic motivation or curiosity [8] [9], on the other hand, is a broader term that can be used to characterize any mechanism that captures *motivation to learn* in an agent [10, 11].

In this paper, we tackle the problem of continual reinforcement learning in a task agnostic setting where the task changes are caused by the non stationarity in the environment. We do so in particular by leveraging the curiosity of the agent to assist itself in knowledge retention.

*1.1. Related Work*

When it comes to solving catastrophic forgetting in the context of reinforcement learning past works have been tried in many ways. the context of continual learning in general early works have focused on using methods that reduce representation overlap [12, 13], replaying samples [14, 15] and using dual architectures [16, 17] with the applications extending to [18]. With the development in the field of neural networks features such as dropout [19] and empirical study on activation functions were also done [20].

Recently the works have focused on handling catastrophic forgetting is via knowledge retention. This can be done by resorting to several methodologies [5], namely, latent parameter storage, distillation, or rehearsals with stored past data in an external memory.

Parameter retention has been done both explicitly and implicitly in several works. For instance, the work in [21] uses a shared latent basis in policy gradient learning in order to capture reusable components of previous policies. Similarly, the contribution in [22] utilizes a shared state-action embedding in order to exploit the shared nature of tasks. Another work based on Progressive Networks [23] uses instead latent representations of the network belonging to previous tasks as an input to the current network. Even though this approach shows immunity to catastrophic forgetting and aids transfer learning, storage requirements grow alongside the amount of tasks. Another promising way of knowledge retention is done by incorporating distillation [24] [25] [26] to retain past knowledge [27, 28, 29].

---

*Corresponding author. Department of Computer Science, University of Maryland College Park, 20742s , USA. Corresponding author email address: pan@umd.edu

In general rehersal based methods [30, 31, 32, 33] has been used to store a limited subset of samples while solving the new task. When it comes to reinforcement learning rehearsal strategies have shown to be an efficient way of retaining knowledge. Rehearsal in RL in general is often associated with offline reinforcement learning methods. In this context, replay buffers have been used as a general form of rehearsal, as opposed to the less commonly used GAN-based rehearsal, also known as *pseudo-rehearsal*. Replay buffers have proven to be sample efficient when it comes to solving MDPs [34] [35] [36]. However, the most commonly used First In First Out (FIFO) buffers become useless when in the context of continual reinforcement learning since, by nature, these replay buffers are designed to discard old transition tuples in favor of the newly encountered ones, resulting in catastrophic forgetting. This warrants a need for the replay buffer to conserve the data corresponding to an older task while collecting data from newer ones. A viable way to do this is by uniformly storing the data that was experienced throughout the agent's lifetime [37].

Some works [38] expand on this idea of a uniformly sampled reservoir by mixing the reservoir with a small FIFO buffer, thus striking for a balance between recent and old samples to attain high performance continual learning. However, a downside is that data is stored proportionally to the time taken by the task itself, thus tasks in which the agent spends more time would get more priority when it comes to storing transition tuples. This can lead to certain tasks that happen in a relatively shorter period getting neglected and as a result, face the risk of catastrophic forgetting. Therefore, the need arises for a buffer to store transitions proportional to the number of tasks faced rather than the amount of time spent on these tasks.

Multi-Time scale Replay buffer [39] on the other hand focuses on maintaining a collection of sub-buffers which are arranged in such a way that the lifetime of a transition tuple in the buffer would follow the power law. This helps the buffer achieve better results over [38] when the change in the environment is not discrete, but happens gradually over time. But inherently even this type of buffer does not retain samples in an equitable manner to all the task regardless on how much the agent is exposed to them.

In reinforcement learning curiosity has been long used as an intrinsic reward (reward generated internally by the agent) to improve an agent's learning process when the extrinsic reward (the reward given by the environment) is sparse thus providing a comparatively more guided learning process. Curiosity has long been used as an umbrella term in the reinforcement learning community as the measure of unexpectedness. This can broadly be categorized as those who measure the novelty of a state [40, 41] or those who measure the uncertainty of the agent in predicting the consequences of its actions [42, 43]. The work in [44] used self-supervised prediction errors as an effective means to compute the curiosity. In particular, they introduced an inverse and a forward model which measures the error in predicting its actions and the next states respectively. This error itself was treated as curiosity to improve the exploration abilities of RL agents. In lines with latter definition curiosity can also be formulated as the error in predicting the reward experienced by an agent. Thus they can be formulated as shown and Fig 1 and named as *inverse*, *forward*, and *reward-based* curiosity.

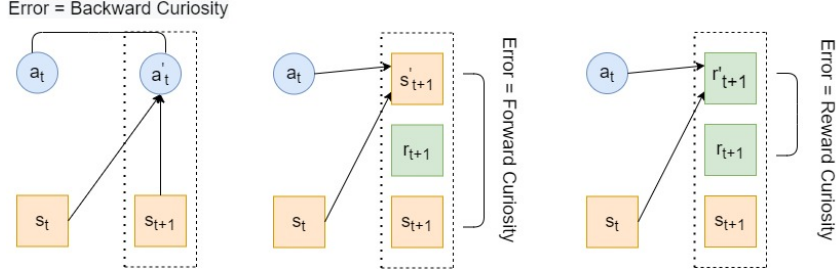

Figure 1: Diagram depicting different types of curiosities that computed as a prediction error associated with predicting certain elements of state action transition namely– backward/inverse curiosity, forward curiosity and reward curiosity. In each of those diagrams $a'_t, s'_{t+1}$ and $r'_{t+1}$ are predicted from $(s_t, s_{t+1}), (s_t, a_t)$ and $(s_t, a_t)$ respectively and the prediction error in them is treated as the curiosity value.

*1.2. Contributions*

In this work, within the continual learning framework paradigm, we primarily limit our focus to stay within the environments where agents are *agnostic to the task changes*. This is one of the most realistic RL scenario happening in real life, as well as in unsupervised and open-ended learning settings [45] and *multi-tasks continual learning* Here we limit the origin of the task change to the changes that may happen in the stationarity of the environment. Throughout the agent's lifetime, these different tasks exist for a different period (certain *tasks are dominant in time*). Within these settings, we aim to answer a fundamental question: *how can we leverage curiosity to improve the composition of replay buffers to improve the performance in continual offline RL?* To this end, we investigate the use of curiosity through the following aspects:

- **Curiosity as a priority metric when it comes to retaining old transition tuples**: Here we try to leverage curiosity's role as a metric of a surprise to prioritize which transitions to store and which ones to discard. This results in a buffer we call the Hybrid Curious Replay Buffer (*HCB*) and they are a combination of both a smaller FIFO buffer and the curious replay buffer. From now on we will identify the buffers that uses a similar combination as hybrid buffers.

- **Curiosity as a tool for task change detection**: Here we use curiosity as a mechanism for task change point detection which, in turn, would allow us to maintain a collection of sub-buffers each of which are dedicated a different task. We call it the Hybrid Reservoir buffer with Task Separation (*HRBTS*).

Primarily, we evaluate these two algorithms against two already existing replay buffer-based solutions [38, 39] that can be used in task agnostic settings. Throughout this work, we will refer to these two existing works as Hybrid Reservoir Buffer (*HRB*) [38] and Multi-Time scale Replay buffer (*MTR*) [39]. On top of the above-mentioned continual learning scope, we also test the proposed algorithms against the settings where task changes in the environment are gradual. This is a setting towards which MTR is aimed at. In summary, we test three different continual learning scenarios – namely, *prolonged task encounter with substantial task drift, and no task re-visitation*, *every timestep task encounter with small task drift and task re-visitation* and *frequent, short-lived task encounter with substantial task drift and task re-visitation*. These tasks are defined and explained further in Section 3.2.

The rest of the paper is organized as follows. Section 2 formulates the types of buffers, Section 3 formally defines the experiments and benchmarks the buffers, and Section 5 concludes and discusses limitations and the future directions for research.

## 2. Methods: Curiosity in Continual Offline Reinforcement Learning

In this section we will motivate and define two different buffers –namely, Hybrid Curious Buffer (HCB) and Hybrid Reservoir Buffer with Task Separation (HRBTS) in Section 2.1 and Section 2.2 respectively.

### 2.1. Using Curiosity as a Priority Measure when Retaining Old Samples

In a continual learning setup when a change in an environment occurs, that change itself can be an indicator of a task change. Even in cases where the change in a task only happens due to a change in the reward function, a reward function-based curiosity can be used to get the same effect. Thus, we formulate the buffer similar to the reservoir buffer [37]. However, in our case curiosity acts as the priority factor to decide whether to retain an old transition tuple, instead of using the time spent on a task as in [37]. This could result in the agent prioritizing the storage of samples from where the environment changed the most, as the curiosity would be higher for those samples. This enables the buffer to keep the transitions that most surprised the agent for a long period, as they are unlikely to be replaced by a transitions that have least surprised the agent. In what follows we will refer to this buffer as *curious buffer*.

In scenarios where the change of task happens abruptly and in short bursts, storing these transitions in the buffer is of paramount importance, as data points are only encountered scarcely. However, on the other hand, if a task is encountered for a long continuous period of time, this buffer can end up prioritizing data points that were initially encountered when the agent was most uncertain about the environment.

### 2.2. Using Curiosity for Task Change Detection

Another way by which we leverage the surprise aspect of curiosity in the context of continual reinforcement learning is by using it as a measure to hypothesize a task change when the tasks are not labeled (i.e. in task-agnostic settings [2]). In essence, we rely on curiosity to draw task boundaries when they are not explicitly given. Here we treat a stream of curiosity data coming from an agent as a signal on which filters from signal processing such as signal to noise ratio filter can be used to detect significant changes, thus, hypothesizing a task change. This task change mechanism is paired with a collection of buffers as defined in Section 2.2.

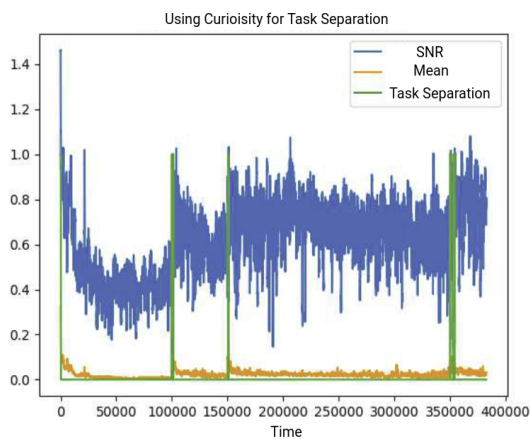

Figure 2: Mean and SNR of curiosity signal $C(t)$ in a continual learning scenario where task changes happen in timesteps [0, 100000, 150000, 350000]. Tasks represent any change that has an influence on either the environment's dynamics or the reward structure. In this case, different tasks corresponds to the change of length in a classic Pendulum environment. The condition $(snr(t) < m_f \cdot \mu(t))$ in combination with eq. 3 and eq. 4 was used to detect the task change where $m$, $l$ and $n$ are hyperparameters tuned for the Hopper environment. Green lines show the points where the condition for $I(t)$ is met and thus, ultimately hypothesize the task boundaries. Here the x axis denotes the time steps while the y axis is a measure of three different quantities, namely, SNR, Mean and Task Separation Indicator.

The filter we use to hypothesize the task boundaries (beginning and ending time instants of the task) using the curiosity measure of the agent in the case of HRBTS buffer is an adaptive threshold filter which is a slight modification of a Signal to Noise ratio (SNR) based filter. The filter is formulated in Section 2.2.1, Expression (4).

*2.2.1. Task Change Detection*

For the sake of notation's clarity, let us refer to the average curiosity of a network at time $t$ as $C(t)$. We maintain a relevant time frame of size $n$, which spans over the most recent $n$ samples of $C(t)$ based on time. That is, the curiosity and its statistics (e.g., mean, variance) at time $t$ are computed based the instantaneous curiosity values recorded from time $t - n$ to time $t$. The mean $\mu(t)$ of the curiosity signal is computed as:

| Notation | Definition |
|---|---|
| $C(t)$ | Curiosity at time $t$ |
| $n$ | Size of the time relevance time frame spanning over curiosity signal |
| $snr(t)$ | Signal to Noise Ratio |
| $\mu(t)$ | Curioisty signal's mean |
| $m_f$ | Hyper parameter pondering the strength of the mean Curiosity C(t) $\mu(t)$ |
| $K(t)$ | Total idle timesteps spent without meeting the selection |
| $k$ | Threshold (hyper parameter) used to determine a hypothetical task change |
| $I(t)$ | Indicator Variable |

Table 1: Terminology and Notations with regards to the adaptive threshold filter in Section 2.2.1

$$\mu(t) = \left(\sum_{t=0}^{n} C(t)\right)/n, \qquad (1)$$

and the signal to noise ratio $snr(t)$ of the curiosity signal is defined as:

$$snr(t) = \mu(t)/(\sigma + \delta), \qquad (2)$$

where $\sigma$ is the standard deviation of the curiosity in the relevant time frame of interval $t - n$ to $t$ and $\delta$ is a small constant that is used to avoid division by zero. Here, $snr(t)$ highlights the points where not only the change in curiosity is higher, but also where the density of changes is largest. A candidate point for task change detection is hypothesized when the boolean condition $snr(t) < m_f.\mu(t)$ is met, where $m_f$ is a hyperparameter. By this condition we filter out points which can be hypothesized as noticeable changes in the curiosity signal.

Normally, we would encounter multiple candidate points around a single task change. These candidate points are further filtered with the help of an idling time hyper-parameter $k$ to ultimately hypothesize a task boundary which is indicated by an indicator variable $I(t)$. This variable is calculated as:

$$K(t) = \begin{cases} K(t) + 1, & \text{if } snr(t) \geq m\mu(t), \\ 0, & \text{otherwise.} \end{cases} \qquad (3)$$

Here $K(t)$ counts the number of timesteps passed since the last task boundary was hypothesized.

$$I(t) = \begin{cases} 1.0, & \text{if } K(t) \geq k \\ 0.0, & \text{otherwise} \end{cases} \qquad (4)$$

Expression (4) effectively tries to avoid multiple task change/task drift point detection in the neighborhood of a single task boundary by the use of a user-specified threshold $k$.

*2.2.2. A Collection of Sub-reservoir Buffers*

Similar to the hybrid reservoir buffer [38], a FIFO buffer smaller than the primary reservoir buffer [37] is maintained along with a reservoir buffer in order to maintain a balance between recent and older data. The reservoir buffer is made up of smaller sub-reservoir buffers of equal size, with each buffer corresponding to a single task as hypothesised by the task boundary detection mechanism. Upon detecting a task boundary space for the sub-reservoir corresponding to the newly detected task is allocated by equally freeing up the space from all the other sub-reservoir buffers. In each of those sub-reservoir buffers, the space is freed up by discarding transitions corresponding to the lowest priority (in the case of the reservoir buffer this corresponds to uniformly discarding samples). Figure 3 illustrates this process in more detail.

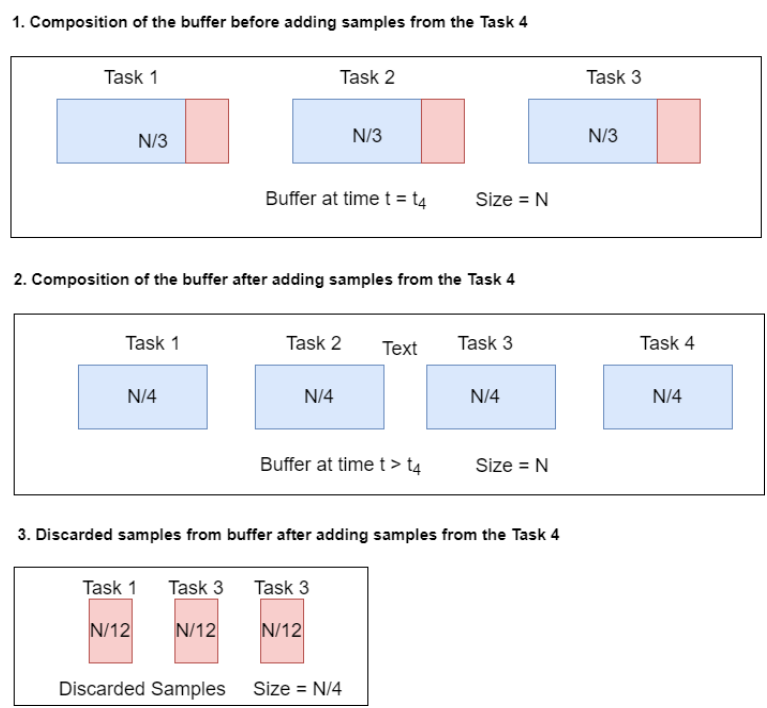

Figure 3: First diagram shows the composition of an HRBTS buffer before a hypothesized task separation at time $t_4$. Here each block corresponds to what the agent thinks as samples belonging to a single task. The red block denotes the samples corresponding to a lower priority measure. Here the size of the buffer is $N$. The second and the third diagrams indicate the composition of an HRBTS buffer after a hypothesized task separation at time $t_4$. The space for the samples from the new task 4 where allocated by discarding the low priority samples (indicated in the third diagram) from the past tasks.

*2.3. Different Types of Buffers using Curiosity*

By using curiosity both as a priority measure and a task change detector we formulate two buffers, namely:

- Hybrid Curious Buffer (**HCB**): A curiosity-based reservoir buffer (as defined in Section 2.1) is paired with a small FIFO buffer.

- Hybrid Reservoir Buffer with Task Separation (**HRBTS**): As defined in Section 2.2, a hybrid buffer is formed by pairing a small FIFO buffer with a collection of sub-reservoir buffers where the number of sub-reservoir was determined by the number of task boundaries that are hypothesized by the filter Eq. 4.

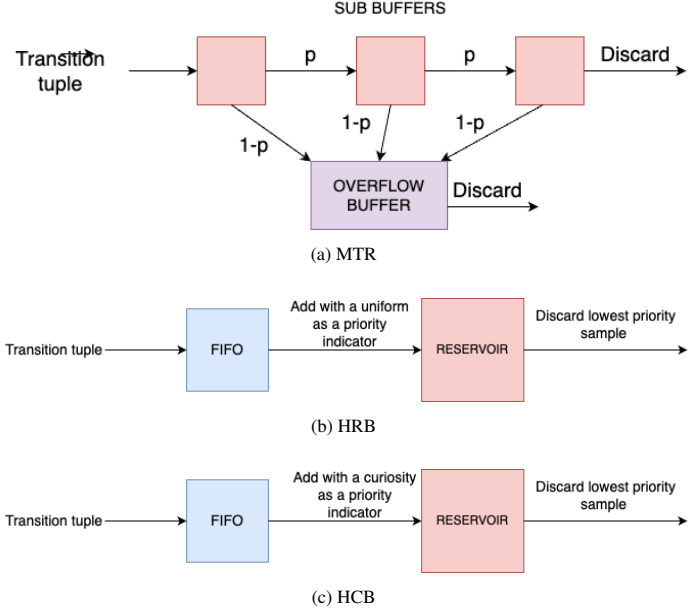

(a) MTR

(b) HRB

(c) HCB

Figure 4: Figure shows the block diagram of MTR, HRB and HCB buffer. In case of MTR the discharged samples from each buffer is pushed into a new buffer with a specific probability until it had to be discarded. In case of HCB and HRB buffer both pushes out samples based on a priority measure (uniform sample in case of HRB and curiosity measure in case of HCB).

## 3. Results: Experiment Setup

Benchmarking environments for continual reinforcement learning is an existing open problem [2]. Past works in the domain of continuous control focused on perturbing certain parameters in the MuJoCo [46] or DM control [47] suite environments. For image-based RL tasks Atari [48] and VizDoom [49] environments have been widely used in the past. Recent works such as [50] have provided an Arcade Learning Environment (ALE) [51] based framework in which the space invaders, breakout and freeway games were modified to provide different difficulty levels in order to benchmark continual and transfer learning environments. In this work similar to the work of [39] we focus on perturbing certain attributes of classic OpenAI Gym [52] and Robo School [53] environments inorder to create a continual learning environment for experimentation. This is further explained in Section 3.2. Furthermore we also consider an experiment based on the MetaWorld framework [54], where different environments were considered as different tasks.

### 3.1. Algorithms

For these above-mentioned tasks Soft Actor Critic (SAC) [36] was used as the base algorithm for learning.

In addition to the first in first out buffer (*FIFO*), we compared the two buffers (HRBTS and HCB) described in Section 2.3 designed using curiosity against Multi-Time scale replay buffer [39] (*MTR*) and a hybrid reservoir buffer [38] (*HRB*). The former emphasizes storing transition tuples such that the lifetime of a single transition tuple in the buffer would, on average, follow the power law. This enables it to emphasize more on the data that is encountered during the agent's intermediate lifetime. The latter, on the other hand, emphasizes storing the transition tuples from different tasks in accordance with the time spent on those respective tasks (uniformly in time).

*3.2. Continual Learning Setup*

We designed a set of experiments to essentially answer three questions:

1. *Question 1*: How well do the proposed curiosity-based algorithms (HRBTS, HCB) hold against existing replay buffer works when an agent's exposure to a certain task in time is non-proportionally higher? This question can further be broken down as follows:

   - How immune can these algorithms be to catastrophic forgetting failure when the agent is exposed to a relatively short task, followed by another time-dominant task that gets a disproportional time of exposure?
   - How much does a time dominant task impedes the subsequent task's learning?

2. *Question 2*: How responsive are algorithms HRBTS and HCB when tasks are encountered in a repetitive and frequent manner?

3. *Question 3*: When tested in settings where MRT [39] is shown to perform best, do the proposed algorithms HRBTS and HCB perform better against MRT which is already shown to outperform HRB [38] in the particular setting?

Experiments were designed to test three different continual learning setups/scenarios in Figure 6, which we believe would best answer the aforementioned questions. To do so, each condition in the environments is formulated by perturbing certain parameters discussed further in Section B. The measurement units of these parameters are based on the definition of the original environment's implementation which in general we would define as in-environment units. These setups/scenarios were broadly categorized into three experimental paradigm settings. Namely:

- *Prolonged task encounter with substantial task drift, and no task re-visitation*: Here during the agent's training time, tasks are encountered only once for a continuous amount of time. Certain tasks in this setting are exposed to the agent for a longer period of time as opposed to others. In practice, a parallel can be drawn between a robot that is asked to learn different tasks in a sequence (without labeling the current task the agent is at) and move on to the other task during the training phase. Thus the agent will not be allowed to revisit the tasks and it has to ultimately generalize in all tasks when it moves on to the testing phase. The experiments in this setting were designed to address Question 1 in two cases which can be summed up by the two sub-questions: **1.** *How immune can the algorithms be to catastrophic failure when a relatively short exposed task is followed by another who gets a disproportional time of exposure to the agent (time dominant)?* and **2.** *How much does a time dominant task impedes the subsequent task's learning?*

- *Frequent, a short-lived task encounter with substantial task drift and task re-visitation*: Here as in the previous case, once a task change happens that tasks last for a certain period of time. However, these changes are frequent and the tasks last for a relatively short period of time in all but one task. Thus, one task remains dominant in terms of exposure time to the agent. As opposed to the prolonged task encounter with substantial task drift, and no task re-visitation, here the same task can be encountered for multiple times. A parallel to this scenario can be drawn to a robot learning to traverse an uneven terrain where it may have to traverse different regions (flat region, mountain areas etc.) of that terrain repeatedly but for a limited amount of time without not having long exposure to that region at every exposure. This was designed to answer Question 2 in Section 3.2. By far, frequent short-lived task encounter with substantial task drift and task re-visiting is the hardest task of the three for several reasons. Here, data from erratic/frequent and short-lived tasks are only encountered for a shorter period of time, which creates several issues:

1. The agent now needs to show an increased amount of urgency in storing these transitions as they are encountered scarcely.

2. Since these tasks only occur for a short period of time, it becomes increasingly difficult for the agent to accurately detect the task boundaries.

3. In these settings it becomes paramount for the agent to maintain some level of knowledge about the previously encountered task if it is to learn efficiently. For instance, if the policy network of the agent maintains no knowledge of the previous task as it encounters it again, then it would start exploring the same initial non-optimal regions and it may never get any data close to the optimal region; thus, resulting in the agent not learning at all at any point.

- *Every timestep task encounter with small task drift and task re-visitation*: This experiment was designed to answer Question 3 in Section 3.2. The settings were designed similar to that in [39] where the tasks change gradually over time. Here, attributes of an environment are constantly and gradually changed over time. In our case, this change follows a sine pattern [39]. The Pendulum's length and Hopper's operational power were changed throughout the agent's lifetime according to Eq. 5 and Eq. 6) (Figures 23 and 25) as:

$$l(t) = l_{min} + \sin(t \cdot 10^{-4}) \cdot (l_{max} - l_{min}), \tag{5}$$
$$p(t) = p_{min} + \sin(t \cdot 10^{-4}) \cdot (p_{max} - p_{min}), \tag{6}$$

where $l_{max}, l_{min}$ correspond to the maximum and minimum values of the length the Pendulum was allowed to take (set to 1.0 and 1.8, respectively, in the environment's units). Similarly, the operating power of the Hopper ($p_t$) was changed at every timestep ($t$) according to Expression (6). $p_{max}, p_{min}$ corresponds to the maximum and minimum values of the power the Hopper was allowed to take (which were 0.75 and 8.75, respectively in the environment's units).

The number of tasks in this setting is equal to the agent's lifetime in timesteps. For evaluation purposes, we analyzed the average rewards corresponding to three tasks that were associated with Pendulum lengths of 1.0, 1.4, and 1.8; and Hopper's operational power of 0.75, 4.75, and 8.75, respectively. Both length and power were measured in the environment's units.

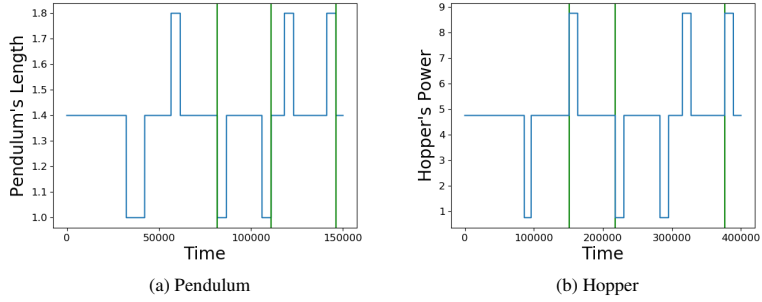

(a) Pendulum

(b) Hopper

Figure 5: Short-lived fluctuations that occurred changing the length (l) in case of Pendulum and power (p) in case of Hopper during the agent's lifetime. Here the y-axis refers to the number of time steps in the agent's lifetime
. In case of the Pendulum, Task corresponding to l = 1.4 was exposed the most to the agent in time whereas, tasks corresponding to l = 1.8 and l = 1.0 occurred frequently at different time scales in short bursts. As per the Hopper's case task corresponding to p = 4.75 was the most exposed in time to the agent whereas, tasks corresponding to p = 0.75 and p = 8.75 occurred frequently at different time scales in short bursts. Here the vertical blue lines correspond to the time frames at which the composition of the buffer was analyzed (as shown in Figure 18).

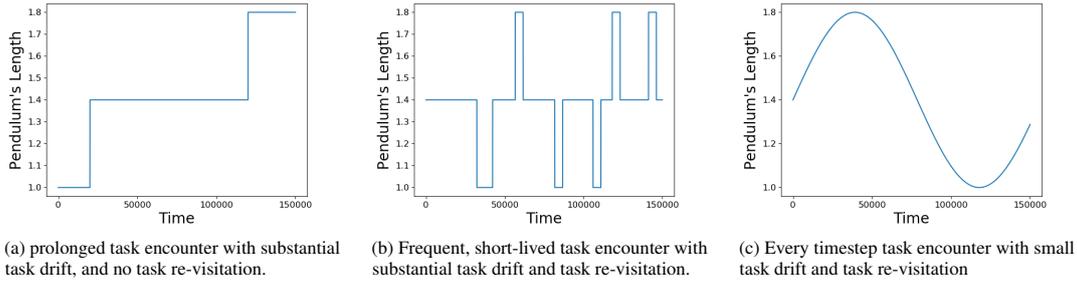

(a) prolonged task encounter with substantial task drift, and no task re-visitation.

(b) Frequent, short-lived task encounter with substantial task drift and task re-visitation.

(c) Every timestep task encounter with small task drift and task re-visitation

Figure 6: These cases indicate a scenario where the length of a Pendulum is changed to mimic the effect of a task change in all three settings. Here the y-axis refers to the number of time steps in the agent's lifetime
. Here each value across the x-axis (Pendulum's length) represents a task (Note that task definitions differ from one another to be adapted to the specific domain of the task. For instance, in Hopper's case, the operational power at which the Hopper operates is changed).

The task drift/ task change in these settings can be further illustrated by Figure 8. Within the scope of this paper task change and task drift are used interchangeably.

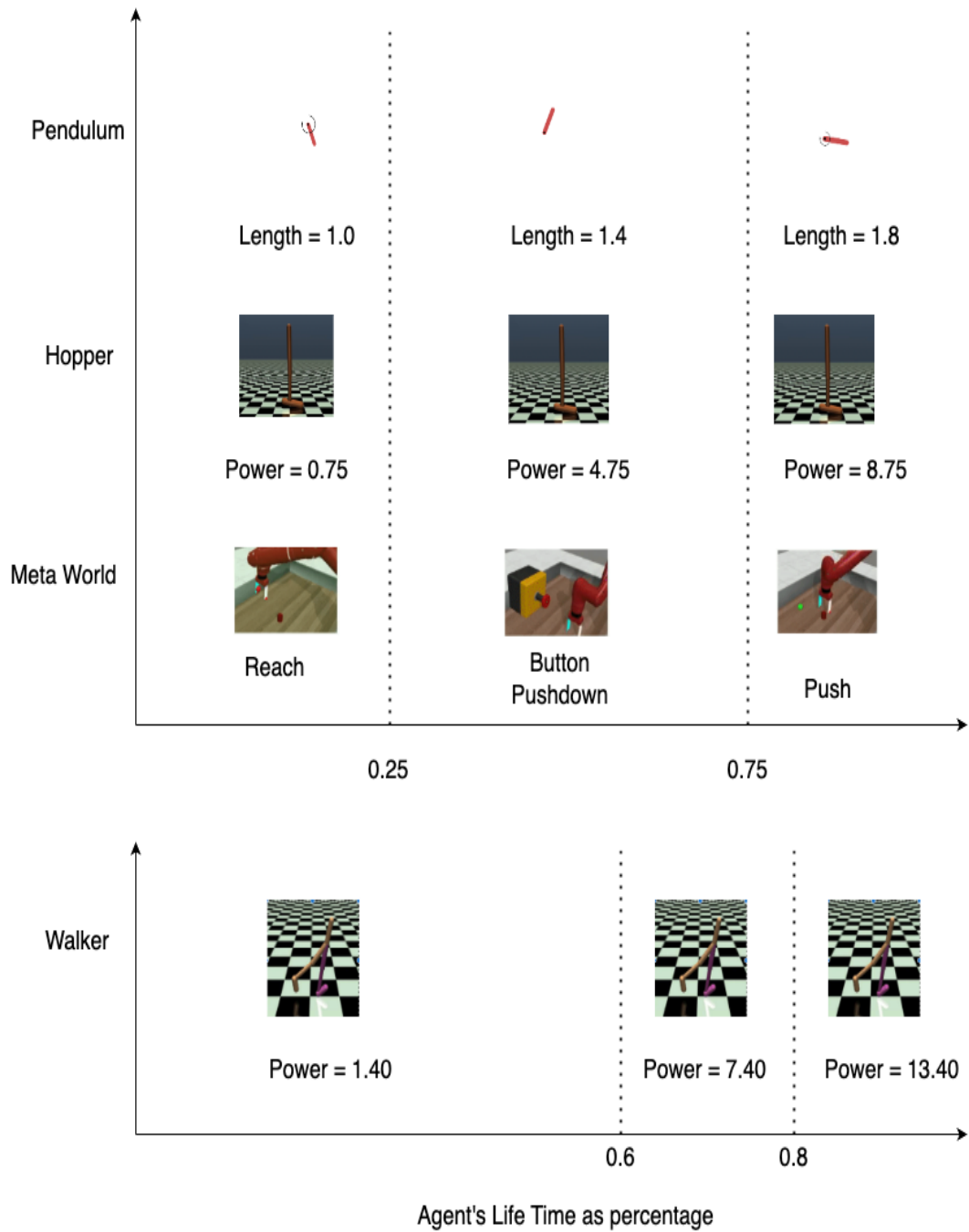

Figure 7: This figure shows the evolution of tasks during the agent's lifetime during the *prolonged task encounter with substantial task drift, and no task re-visitation* setting. Here the y axis refers to the percentage of agent's total life time and where the task change happens.

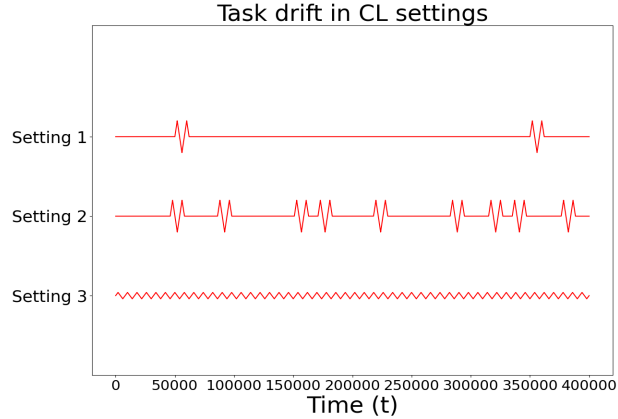

Figure 8: The figure (extended from [2] nomenclature) shows different ways in which each new task is encountered according to the nature of the task drift (TD) defining the task change in our experiments. Here the y-axis refers to the number of time steps in the agent's lifetime
. Here the small task drift and substantial task drift as noted in the diagram corresponds to how different the newer tasks are in comparison to the previous one. *Setting 1*: (prolonged task encounter with substantial task drift, and no task re-visitation), new task change happens in a non-periodic manner with certain tasks lasting longer than the others. Furthermore, the amount of task change or TD that occurs in the new task compared to the previous one is higher than in setting 3. *Setting 2*: (Frequent, short-lived task encounter with substantial task drift and task revisitation) only differs from setting 1 in terms of how frequent the task change happens. *Setting 3*: (frequent, every timestep task encounter with a small TD) Although the TD is smaller compared to the other two settings, it happens at every timestep (periodic).

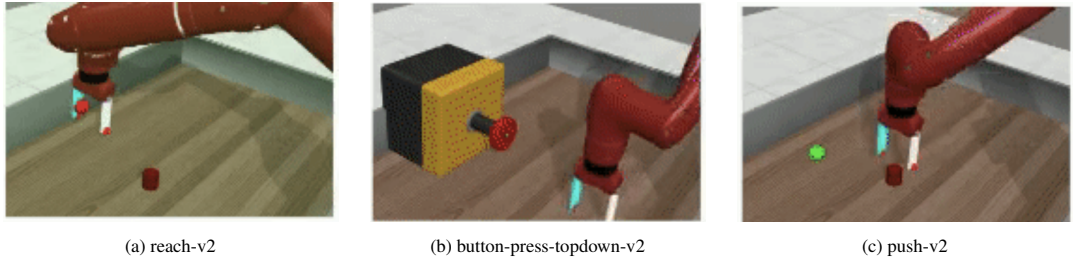

(a) reach-v2  (b) button-press-topdown-v2  (c) push-v2

Figure 9: Three tasks chosen from the Metaworld environment.

*3.3. RL models architecture*

In order to be in line with the works in comparison [39], [38] we choose Soft Actor Critic [36] as our base reinforcement learning algorithm with the following architecture. All settings used a learning rate of 0.0003 (similar to the learning rate selected by the SAC paper [36]). Pendulum, Hopper, and Walker 2D used a feed-forward network with two hidden layers of 256 hidden units each and a batch size of 512.

*3.4. Architecture for curiosity-guided strategies*

In all experiments 5% of the buffer's capacity was allocated for the FIFO buffer whenever it was used alongside the other buffers in order to prioritize the use of recent data on some level. This was used to balance newer and older experiences. A separate FIFO buffer was used to train the curiosity network.

# 4. Results: Experiments and Analysis

*4.1. Results and Analysis: prolonged task encounter with substantial task drift, and no task re-visitation*

Experiments in Pendulum and Hopper and MetaWorld were designed to answer the former while the experiment in the Walker environment was designed to answer the latter. In the case of Hopper and Pendulum, an initial relatively short-lived task was followed by a time dominant intermediate task. In the Walker's case, an initial time-dominant task was followed by two subsequent relatively short-lived tasks. Fig 10, Fig 11 and Fig 16 denote the composition of the replay buffer at different time steps. When there is no additional information about the correlation between tasks ideally we hypothesize the best replay buffer to store an equal amount of transitions corresponding to each task the agent had encountered so far regardless of how long the agent was exposed to a single task. Rest of the figures in the section plots the performance of the agent in each task (how capable is the agent in solving the task) during it's lifetime.

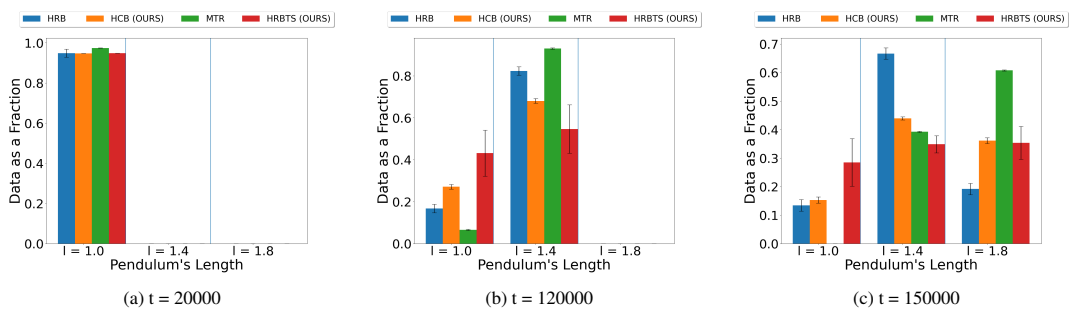

Figure 10: *Pendulum* (prolonged task encounter with substantial task drift, and no task re-visitation): Vertical bars indicate the amount of data belonging to each task as a ratio of the total buffer size at different time frames. Vertical blue lines partition the plot's space for separate tasks. Results were averaged over 8 runs. Here HBTS (ours) (Section 2.2) maintains a better representation for all tasks compared to HCB (ours) (Section 2.1), HRB (Section 2.2) and MTR [39]. Out of the four, MTR buffer is the most biased towards the most exposed task in time, l = 1.4, when it comes to storing data.

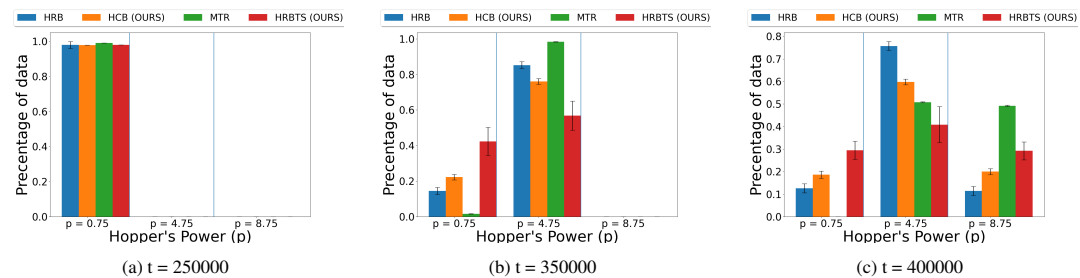

Figure 11: *Hopper* (prolonged task encounter with substantial task drift, and no task re-visitation): Vertical bars indicate the amount of data belonging to each task as a ratio of the total buffer size at different time frames. Vertical blue lines are used to partition the graph's space for separate tasks. Here the vertical blue lines marks the start of the subsequent task (on the right) and the end of the previous task (on the left). Results were averaged over 8 runs. Tasks were characterized by different driving power p of Hopper, measured in environment's units of the Hopper in the environment. Here p = 4.75 is the setting representing the time dominant task. HBTS (ours) maintains a better balance between the data corresponding to all tasks in the buffer compared to HCB (ours), HRB [38] and MTR [39].

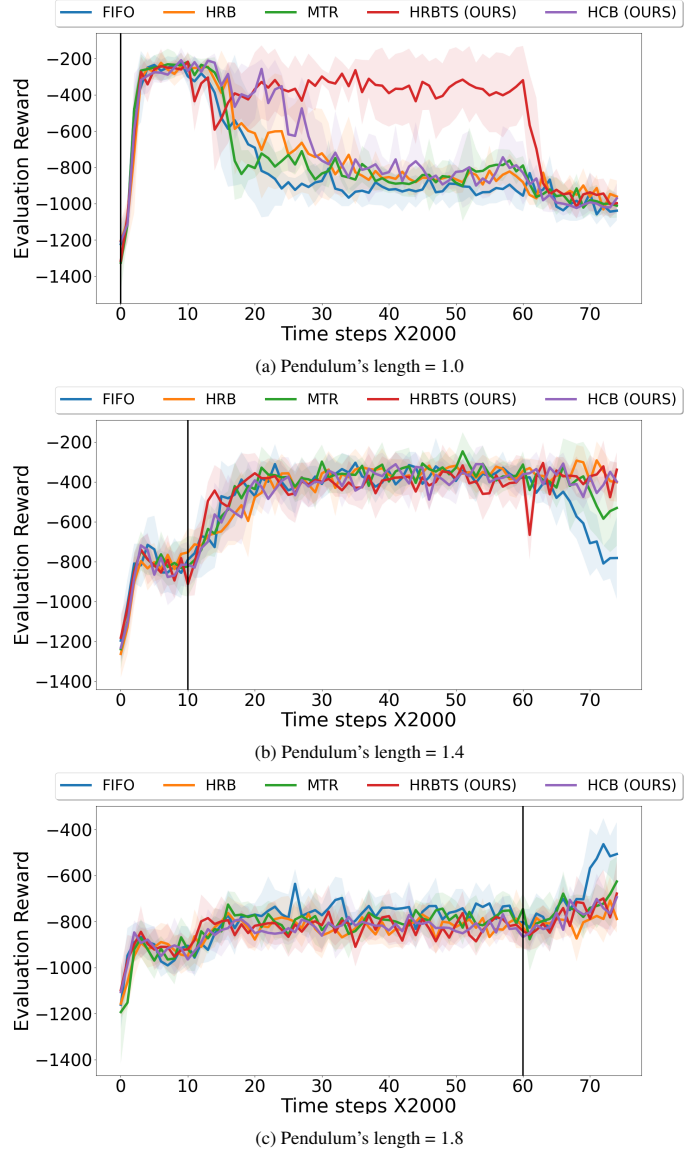

Figure 12: Prolonged task encounter with substantial task drift, and no task re-visiting - *Pendulum*: Individual task's evaluation rewards corresponding to FIFO buffer, HRB [38], MTR [39], HRBTS (ours) (Section 2.2) HCB (ours) (Section 2.1). Rewards were averaged over 8 runs. Black vertical lines denote the actual task change (where the length l (measured as defined in the environment) of the Pendulum was changed). This experiment tests an algorithm's ability to remember short-lived tasks when they are followed by a subsequent time dominant task (corresponding to Pendulum length l = 1.4). The short-lived initial task corresponding to l = 1.0 HRBTS (ours) displays the least tendency to forget Even though HCB (ours) was immune to catastrophic forgetting to some extent, it eventually falls short of HRBTS (ours). FIFO tends to forget more. It even forgets the most exposed task in time, for length l = 1.4, almost immediately. FIFO had shown a better ability to learn new tasks at the cost of forgetting old ones.

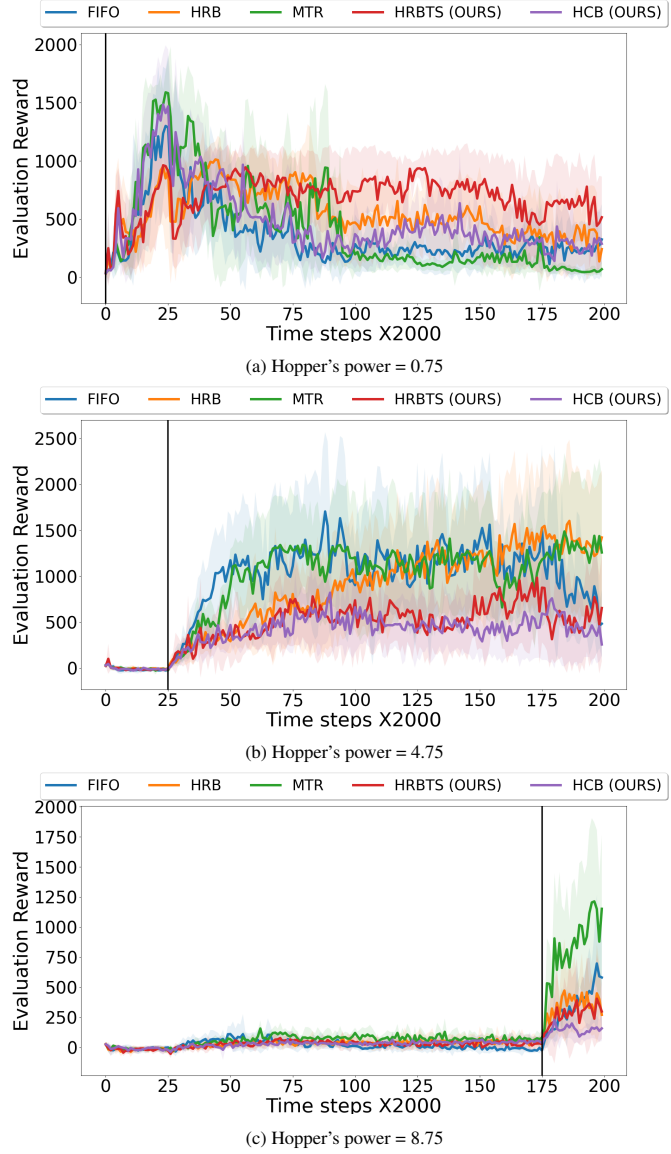

Figure 13: Prolonged task encounter with substantial task drift, and no task re-visiting - *Hopper*: Individual task's evaluation rewards corresponding to FIFO buffer, HRB [38], MTR [39], HRBTS (ours), HCB (ours) were plotted . Rewards were averaged over 8 runs. Black vertical lines denote the actual task change (where the corresponding driving power p, measured in environment's units, of the Hopper was changed). This experiment tests an algorithm's ability to remember short-lived tasks when they are followed by a subsequent time dominant task. power = 4.75 is the most exposed task in time. The short-lived initial task corresponding to power = 0.75, HRBTS (ours) displays the least tendency to forget. FIFO and MTR show tend to forget more. FIFO and MTR had shown a better ability to learn new tasks at the cost of forgetting old ones.

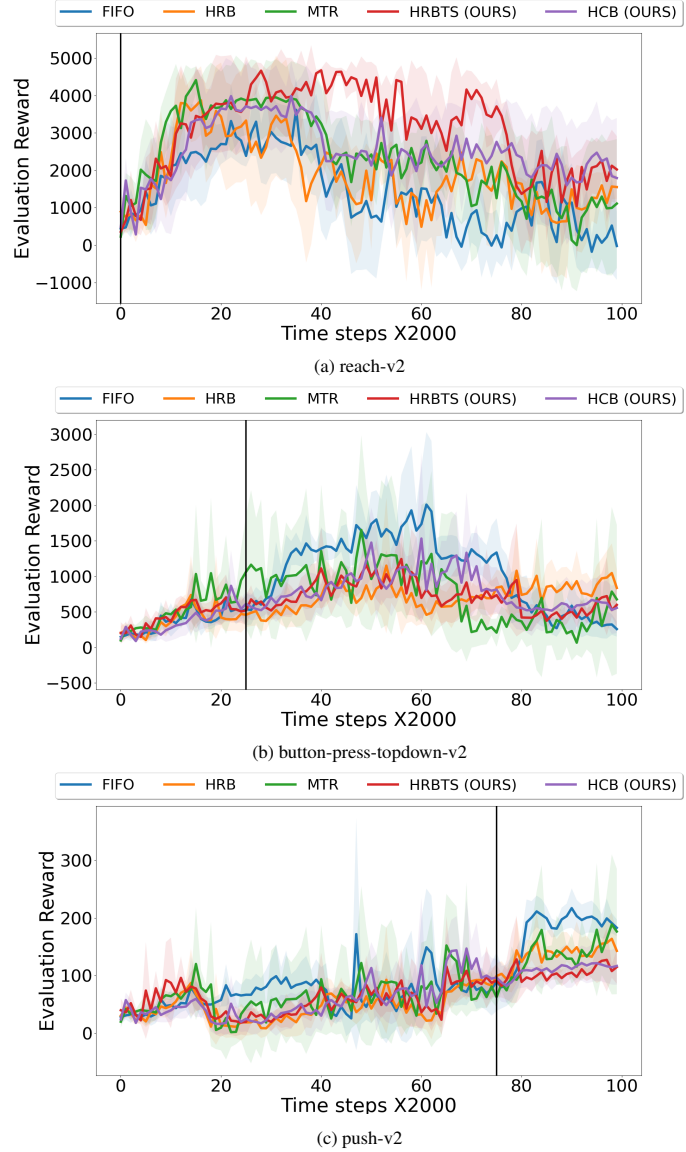

Figure 14: Prolonged task encounter with substantial task drift, and no task re-visiting - *Metaworld*: Individual task's evaluation rewards corresponding to FIFO buffer, HRB [38], MTR [39], HRBTS (ours), HCB (ours) were plotted . Rewards were averaged over 8 runs. Black vertical lines denote the actual task change . This experiment tests an algorithm's ability to remember short-lived tasks when they are followed by a subsequent time dominant task. button-press-topdown-v2 is the most exposed task in time. The short-lived initial task corresponding to reach-v2. HRBTS (ours) displays the least tendency to forget. FIFO and MTR show tend to forget more. FIFO and MTR had shown a better ability to learn new tasks at the cost of forgetting old ones.

The results for all of the algorithms in the first two experiments can be summarized as follows.

1. **HRBTS**: As seen in Figure 10 and Figure 11, HRBTS was able to take advantage of task separation labels and was able to maintain an even composition of data in the buffer for all tasks that were encountered. It eventually had led to HRBTS performing better when it comes to remembering the shortly exposed initial task.

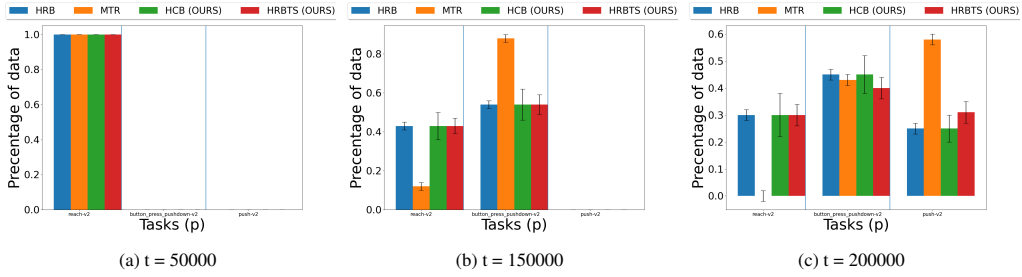

(a) t = 50000  (b) t = 150000  (c) t = 200000

Figure 15: *Meta World* (prolonged task encounter with substantial task drift, and no task re-visitation): Vertical bars indicate the amount of data belonging to each task as a ratio of the total buffer size at different time frames. Vertical blue lines are used to partition the graph's space for separate tasks. Here the vertical blue lines marks the start of the subsequent task (on the right) and the end of the previous task (on the left). Results were averaged over 8 runs. button-press-topdown-v2 is the most exposed task in time. The short-lived initial task corresponding to reach-v2. HRBTS (ours) maintains a better balance between the data corresponding to all tasks in the buffer compared to HCB (ours), HRB [38] and MTR [39].

2. **HCB**: HCB also maintains a better composition (in terms of transition data corresponding to different tasks) but it falls behind HRBTS. Though HCB had shown to perform better than most algorithms when it comes to remembering the shortly exposed task it eventually falls behind HRBTS in terms of being immune to catastrophic forgetting. This shows one of the pitfalls of using curiosity alone as a priority metric. Curiosity is a volatile quantity that can change unpredictably across tasks. This can result in it not being able to capture data from all the regions of the environment.

3. **FIFO and MTR**: Both FIFO and MTR had shown somewhat of similar behavior with MTR being somewhat conservative when it comes to discarding old samples compared to FIFO. Still, both of them weren't ultimately able to maintain an even composition of samples from all tasks which had led to them failing to perform better than both HRBTS and HCB.

4. **HRB**: As seen in Figure 10 and Figure 11 HRB favoured task that was long exposed to the agent and thus resulted in performing better at that task while not being able to adapt to older or newer short-lived tasks.

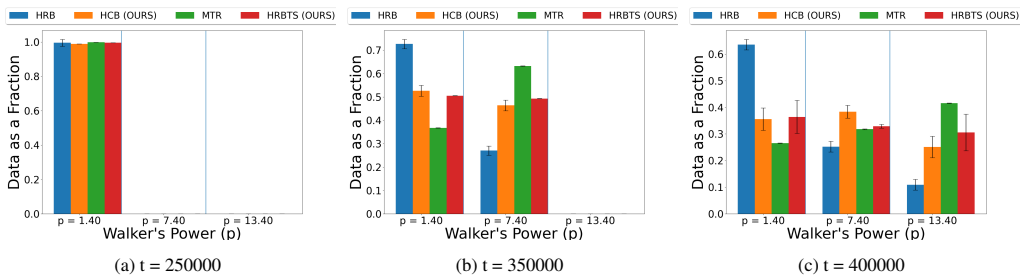

(a) t = 250000  (b) t = 350000  (c) t = 400000

Figure 16: *Walker* (prolonged task encounter with substantial task drift, and no task re-visitation): Vertical bars indicate the amount of data belonging to each task as a ratio of the total buffer size at different time frames. Vertical blue lines are used to partition the graph's space for separate tasks. Here the vertical blue lines marks the start of the subsequent task (on the right) and the end of the previous task (on the left). Results were averaged over 8 runs. Tasks were characterized by different driving powers (p, measured in environment's units) of the Walker in the environment. Here p = 7.4 is the time dominant task. Here HRBTS (ours) (Section 2.2) maintains a better composition of data for all tasks in the buffer compared to HCB (ours) (Section 2.1), HRB (Section 2.2) and MTR [39].

As mentioned before the third experiment was primarily done to check how much does a time-dominant task impedes the subsequent task's learning? The results can be summarized as below.

1. **HRBTS and HCB**: As with the previous case both HRBTS (Section 2.2) and HCB (Section 2.1) where able to maintain better composition of samples in the replay buffer for all tasks as seen in Figure 16. Though HRBTS wasn't as good as FIFO and MTR when it comes to adapting to newer tasks it did adopt better than HRB. Though HCB did well in the third task it failed to do so in the second again highlighting the volatility of using curiosity as a priority metric.

2. **FIFO and MTR**: FIFO buffer was the least impeded by the previous time dominant task while learning the subsequent. This can be attributed to it's the ability to forget past tasks almost immediately due to it's first in first out nature when it comes to retaining the samples. MTR [39] came in second to FIFO. Even though MRT don't discard the past transition tuples as immediate as the FIFO buffer does, because it preserves the samples based on power law, it does forget the past samples faster than HRBTS (Section 2.2), HCB (Section 2.1) and HRB [38] as we can see in Figure 16.

3. **HRB**: As in the previous experiments HRB was partial towards the most exposed task to the agent as a result it performed the lead when it comes to newer short-lived tasks.

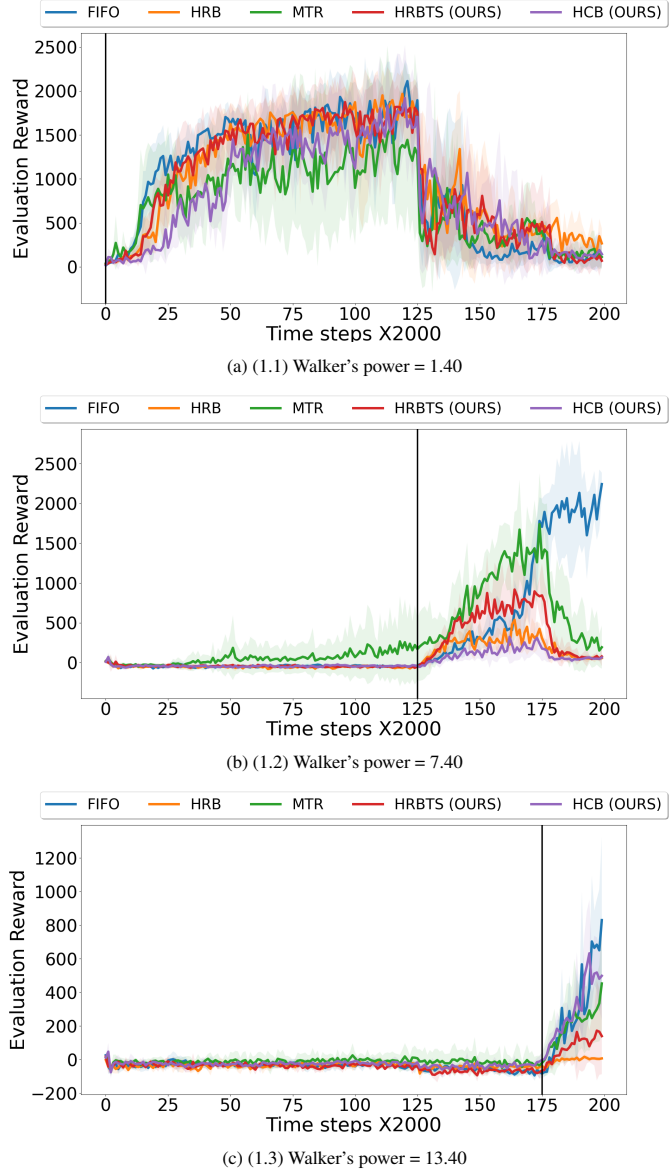

Figure 17: *Walker2D* (prolonged task encounter with substantial task drift, and no task re-visitation): Individual task's evaluation rewards corresponding to FIFO buffer, HRB [38], MTR [39], HRBTS (ours) (Section 2.2) HCB (ours) (Section 2.1) . Here the rewards were averaged over 8 runs. Black vertical lines denote the actual task change (where the driving power (p, measured in environment's units) of the Walker was changed). This experiment tests the algorithm's ability to learn newer short-lived tasks effectively when they are followed by a time dominant task. The initial task corresponding to power = 1.4 is the most exposed task in time. When it comes to learning the subsequent tasks (corresponding to power = 7.4 and power = 13.4 ), FIFO's learning was the least affected by the time dominant initial task. MTR [39] comes in second when it comes to learning these subsequent tasks. HRBTS (ours) (Section 2.2) fall in third . Even though HCB (ours) (Section 2.1) was able to learn the task corresponding to power = 13.4 effectively, it fails relative to the others when it comes to the intermediate task corresponding to power = 7.4. HRB's [38] learning was affected the most by the time dominant task.

*4.2. Results and Analysis: Frequent, short-lived task encounter with substantial task drift and task re-visitation*

Fig 18 and Fig 19 denote the composition of the replay buffer at different time steps. When there is no additional information about the correlation between tasks ideally we hypothesize the best replay buffer to store an equal amount of transitions corresponding to each task the agent had encountered so far regardless of how long the agent was exposed to a single task. Rest of the figures in the section plots the performance of the agent in each task (how capable is the agent in solving the task) during it's lifetime.

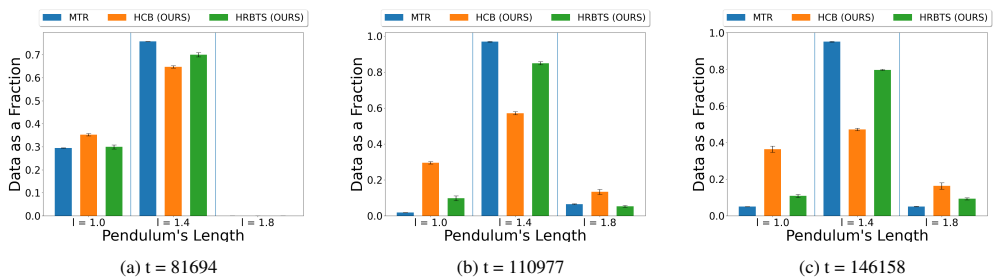

(a) t = 81694   (b) t = 110977   (c) t = 146158

Figure 18: *Pendulum* (Frequent, short-lived task encounter with substantial task drift and task re-visitation): Vertical bars indicate the amount of data belonging to each task as a ratio of the total buffer size at different time frames. Vertical blue lines are used to partition the graph's space for separate tasks. Here the vertical blue lines marks the start of the subsequent task (on the right) and the end of the previous task (on the left). Results were averaged over 8 runs. Here HCB (ours) (Section 2.1) maintains a better representation for all tasks compared to HRBTS (ours, Section 2.2) and MTR [39]. Out of the three MTR buffer is the most biased towards the most exposed task in time, l = 1.4, when it comes to storing data.

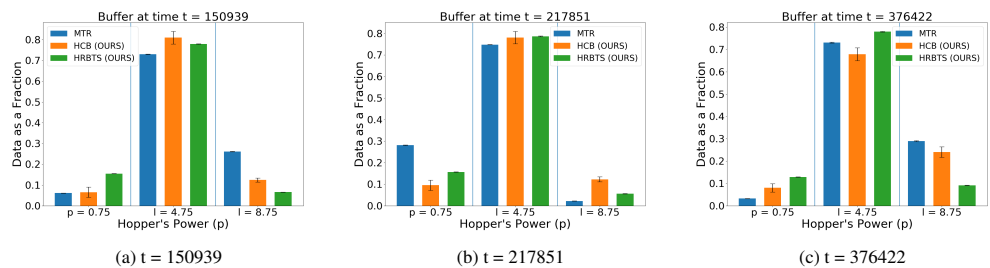

(a) t = 150939   (b) t = 217851   (c) t = 376422

Figure 19: *Hopper* (Frequent, short-lived task encounter with substantial task drift and task re-visitation): Vertical bars indicate the amount of data belonging to each task as a ratio of the total buffer size at different time frames. Vertical blue lines are used to partition the graph's space for separate tasks. Here the vertical blue lines marks the start of the subsequent task (on the right) and the end of the previous task (on the left). Results were averaged over 8 runs. Task corresponding to Hopper's operating power = 4.75 is the time dominant task. Unlike in the Pendulum's case here, HCB (ours) (Section 2.1) couldn't maintain an even composition in the buffer showcasing the volatile nature of the curiosity signal. However, it along with HRBTS (ours) Section 2.2 was consistent in its composition as opposed to MTR [39] whose composition was reactive towards changes. That is when a task change happens it allocated higher storage space in the buffer for the task but it discards those samples immediately in favor of the next task change as seen in figures corresponding to t = 217851 and t = 376422.

- **HRBTS and HRB** : Due to the erratic and short-lived nature of the tasks changes it became harder for the task separation mechanism 2.2 to accurately detect the task changes which was reflected in composition of the buffer as seen in Figure 18, Figure Figure 19. In the Pendulum's case, HRBTS (Section 2.2) still managed to maintain more data from the short-lived tasks compared to MTR buffer [39]. This resulted in HRBTS performing better at remembering the task corresponding to Pendulum length (l) = 1.0. Though in the Pendulum's case the filter was able to detect some changes at least, in the Hopper's case it wasn't able to detect any leading to the HRBTS (Section 2.2) acting similar to HRB [38]. One disadvantage HRBTS had in this setting was, even if it managed to capture all the task boundaries due to its inability to differentiate across tasks it would have ended up allocating several sub-buffers for a single task; thus, becoming partial to that one task. This could also have been a reason why HRBTS trailed HCB here in terms of performance.

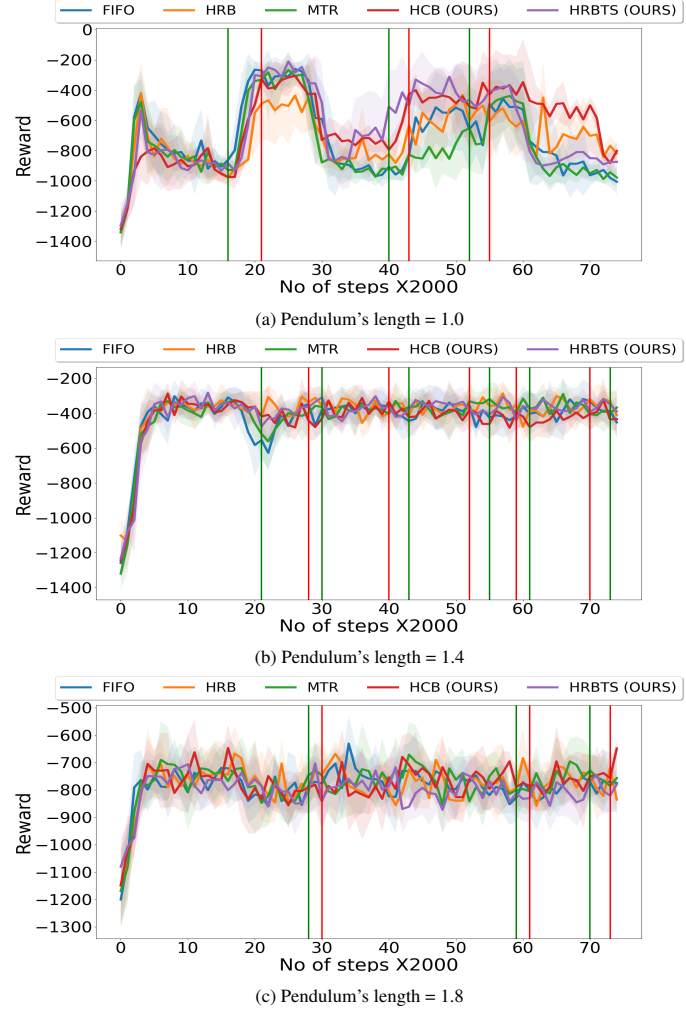

(a) Pendulum's length = 1.0

(b) Pendulum's length = 1.4

(c) Pendulum's length = 1.8

Figure 20: Frequent, short-lived task encounter with substantial task drift and task re-visitation - *Pendulum*: Here the green vertical lines indicate the start of the corresponding short-lived tasks and red lines indicate the end of those tasks (according to Figure 5). For instance, in the first diagram green line indicate the start of the task corresponding to Pendulum's length = 0.75 while red indicates the end of it. Individual task's evaluation rewards corresponding to FIFO buffer, HRB [38], MTR [39], HRBTS (ours) (Section 2.2) HCB (ours) (Section 2.1) . Here the rewards were averaged over 8 runs. Black vertical lines denote the actual task change (where the length (l, measured in environment's units) of the Pendulum was changed). This experiment tests an algorithm's ability to remember a task that is encountered frequently in short bursts. Task corresponding to l = 1.4 is the most exposed task in time. All algorithms performed equally on the task corresponding to l = 1.8. When it comes to the task corresponding to l = 1.0 HCB (ours) learns the best which was followed by HRBTS (ours). HRB comes in third. MTR and FIFO performed equally.

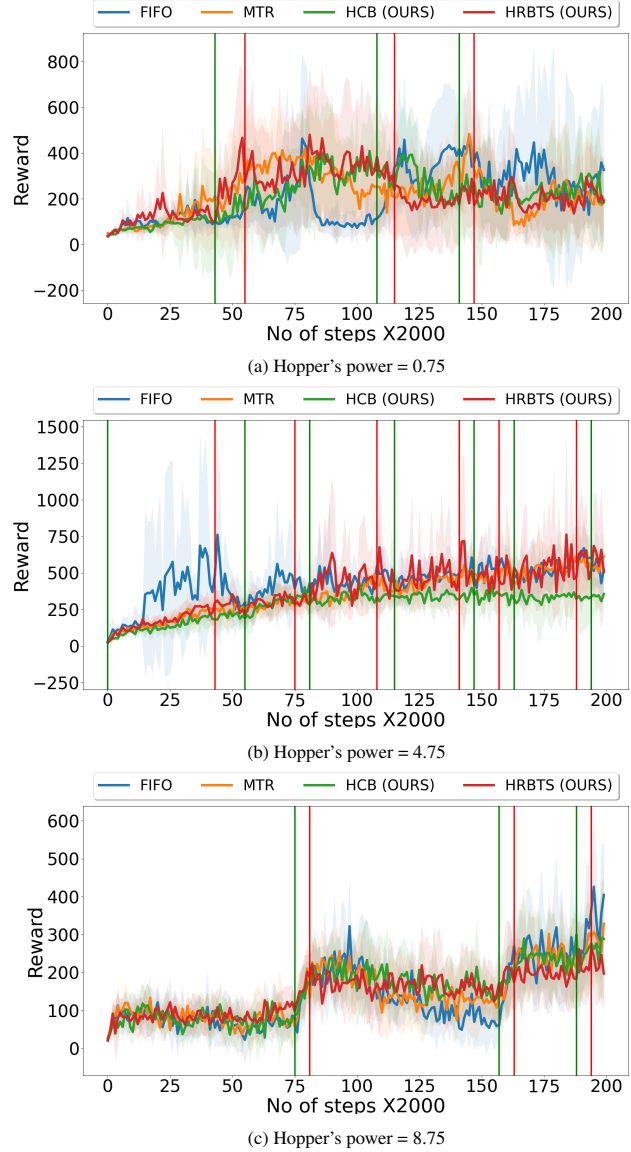

Figure 21: Frequent, short-lived task encounter with substantial task drift and task re-visitation - *Hopper*: Individual task's evaluation rewards corresponding to FIFO buffer, HRB [38], MTR [39], HRBTS (ours) (Section 2.2) HCB (ours) (Section 2.1) . Here the rewards were averaged over 8 runs. Here the green vertical lines indicate the start of the corresponding short-lived tasks and red lines indicate the end of those tasks (according to Figure 5). For instance, in the first diagram green line indicate the start of the task corresponding to Hopper's operational power (p) = 0.75 while red indicates the end of it. This experiment tests an algorithm's ability to remember a task that is encountered frequently in short bursts. Task corresponding to p = 4.75 is the most exposed task in time. FIFO seemed the most reactive to the changes but it also showed tenancy to forget sooner as seen in the figure corresponding to power = 0.75. As the agent is often exposed to all tasks all the other buffers managed to learn those tasks. Even though as seen in Figure 19 HCB (ours) (Section 2.1) and HRBTS (ours) (Section 2.1) was able to have a consistent composition in the buffer they were not enough to make significant gains in terms of generalization as in the Pendulum's case.

- **HCB** (Section 2.1): Interestingly, HCB (Section 2.1) was able to capture much data from the recurring tasks and was able to maintain a more even composition of tasks in the buffer at all times. This highlights the potential of curiosity as a priority measure. HCB had shown to possess the ability to show urgency in storing samples from tasks they are scarce. This had resulted in HCB outperforming the rest in terms of catastrophic forgetting (in task corresponding to length = 1.0). However, when it comes to Hopper HCB wasn't able to maintain as much of an even composition as it was able to in the Pendulum's case and it was not enough to net any significant gains for HCB in terms of catastrophic forgetting. Though HCB has the potential to be urgent with regards to sudden changes, this showcases its weakness which is its dependence on a volatile quantity like curiosity.

- **MRT and FIFO** Both MRT [39] and FIFO in these cases performed similarly. They weren't intelligent about storing enough samples from the short-lived erratic tasks which resulted in them not being able to show much of immunity to catastrophic forgetting in any of the experiments.

All buffers performed equally in terms of the task that the agents where the most exposed to.(length = 1.4).

### 4.3. Results and Analysis: Every timestep task encounter with small task drift and task re-visiting

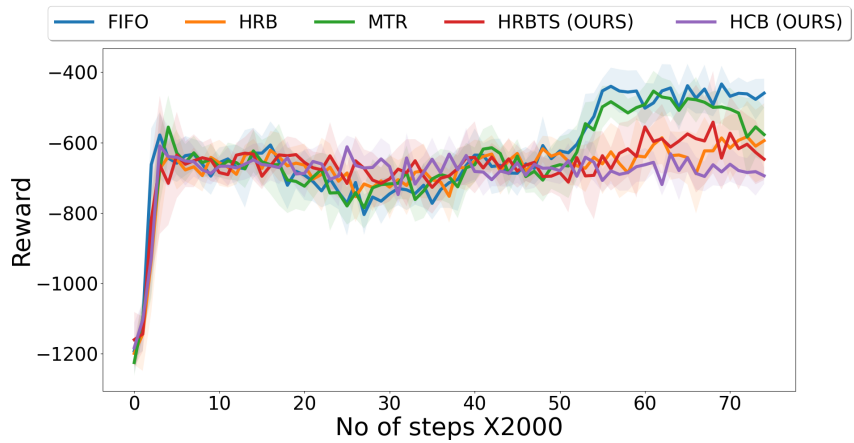

Figure 22: Pendulum's length = 1.0.

Figure 23: *Pendulum* (Every timestep task encounter with small task drift and task re-visitation, where fluctuations follow a sine pattern: An average of the evaluation reward corresponding to FIFO, HRB [38], MTR [39], HCB (ours) (Section 2.1) and HRBTS (ours) (Section 2.2). Here the average was taken across three tasks corresponding to the Pendulum's length (l, measured in environment's units) of 1.0, 1.4, and 1.8. The purpose of this experiment is to check how sensitive is the algorithm when the environment around it evolves constantly. MTR along with FIFO on average performed better but less robust.

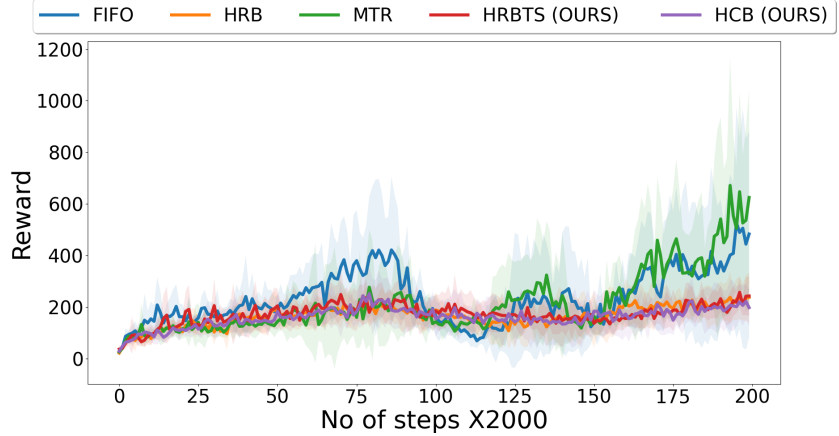

Figure 24: Hopper's power = 4.75

Figure 25: *Hopper* (Every timestep task encounter with small task drift and task re-visitation, where fluctuations follow a sine pattern): An average of the evaluation reward corresponding to FIFO, HRB [38], MTR [39], HCB (ours) (Section 2.1) and HRBTS (ours) (Section 2.2). Here the average was taken across three tasks corresponding to the Hopper's driving power (p, measured in environment's units) of 0.75, 4.75, and 8.75. The purpose of this experiment is to check how sensitive is the algorithm when the environment around it evolves constantly. MTR along with FIFO on average performed better, but less robust.

The results of this experiment in Fig 23 and Fig 25 can be summarized as follows:

1. **HCB, HRBTS, and HRB**: Since in this scenario, the task change essentially happened at every timestep, curiosity was not as significant as in previous scenarios. As the agent faced a new environment at every timestep, the agent was denied of a sudden surprise, as opposed to scenarios where task changes happen after some time after the last task change. Thus, HRBTS failed to detect any task boundaries and ended up behaving similar to HRB. Similarly, as the agent was surprised by a new environment at every timestep, it ended up acting similarly to HRB. All three of them fell behind MTR [39] and FIFO when it comes to the average reward.

2. **FIFO and MTR**: Both in the Hopper and Pendulum environments FIFO and MTR were able to show some kind of tendency to generalize for tasks at a short-range. The number of tasks in this setting is equal to the agent's lifetime in timesteps. For evaluation purposes, we analyzed the average rewards corresponding to three tasks that were associated with Pendulum lengths at 1.0, 1.4, and 1.8; and Hopper's operational power at 0.75, 4.75, and 8.75 respectively. Both length and power were measured in the environment's units.

This experiment showcased the weakness in both algorithms that depend on curiosity signal when it comes to changes in the task id at every timestep.

*4.4. Ablation study*

The addition and removal of certain features from the proposed buffers would make them equivalent to certain other buffers against which they are benchmarked.

- When it comes to the hybrid reservoir buffer with task separation (HRBTS) if we are to remove the Signal to Noise ratio-based task separation mechanism, the resulting buffer would, in practice be equivalent to a hybrid reservoir buffer (HRB).

- In the case of hybrid curious replay buffer (HCB), if we do not prioritize the samples with curiosity in practice it would result in the buffer performing equivalently to the first in first out replay buffer (FIFO).

# 5. Conclusion and Future Work

In this work, we investigate the ways to use curiosity on replay buffers to improve performance in a multi-task, task agnostic offline continual reinforcement learning setting where certain tasks are exposed to the agent for a longer period of time than the others. We use curiosity both as a priority metric to fill up the buffer and manage its content and as a task boundary detector.

Using each of these curiosity signals, we create two buffers, namely hybrid curious replay buffer (HCB) and a hybrid reservoir buffer with task separation (HRBTS). We test them against two replay buffer-based methods that are known for task agnostic offline continual reinforcement learning: Multi-Time scale replay buffer (MTR) [39] and Hybrid Reservoir Buffer [38]. In most settings, both proposed buffers showcase better ability to resist catastrophic forgetting compared to the others.

When the tasks are encountered for a prolonged period of time with a substantial task drift/task change, HCB (Section 2.1) falls behind HRBTS (Section 2.2) in terms of being immune to catastrophic forgetting. However, it provides a better result than the rest, including HRBTS, when tasks appear frequently with a substantial task drift but for a shorter period of time. Even though, on its own HCB does not perform ideally, it does open up scope for it to be used in hybrid with some other buffers due to its ability to prioritize storage when tasks happen frequently with a shorter life span. However when tested against MTR on the setting where the changes in the environment are not sudden (smaller task drift) but occur at every timestep

When it comes to the suitability of the algorithms towards open-world learning (where new tasks emerge that have not been encountered yet), the current architecture does not explicitly have an architectural structure to facility the learning. Since the core of the replay buffer's design depends on storing samples in an equitable way among the tasks, any algorithm that facilitates the generalization given an equal number of data samples among tasks can make better use of our replay buffers when compared to using traditional replay buffers to get better results.

*5.1. Future Works*

As far as future works are concerned, there are some meaningful future directions that can be explored with regards to both the proposed buffers. First, the HRBTS buffer in its form requires a large amount of hyper-parameter tuning in order to find accurate task boundaries. Some research needs to be done on having a task separator with a lesser number of hyperparameters to tune. Secondly, even though this buffer is reactive to task boundaries, it has no knowledge of task similarity, and thus, when a single task is encountered multiple times it would result in allocating multiple sub-buffers for a single task; thus, favoring that task over the others. Therefore, another meaningful research could be devising ways to formulate task similarities in a principle manner, so that the above-mentioned problem would be avoided.

When it comes to HCB, even though it is not the best approach across all tasks, the HCB approach shows some urgency in retaining knowledge from very short-lived tasks that other approaches do not. This opens up a potential for improvement in HCB in two ways. First, it can be used in hybrid with other buffers as a component for capturing erratic surprises to capture the best of both worlds. Secondly, a persisting issue with HCB is that it only prioritizes surprise. Furthermore, as seen in the prolonged task encounter with substantial task drift and no task re-visiting, this can lead to the buffer solely prioritizing earlier transitions because due to the larger prediction error (as a result those transitions will have a higher curiosity) while ignoring the subsequent data where the prediction error associated to curiosity is much smaller and thus, result in the agent not learning effectively. In consequence, a potential solution to be explored to solve this issue is to use a collectio of different sub-curious buffers, each of which would be responsible for capturing transitions corresponding to curiosity in different value ranges.

**Compliance with Ethical Standards**

- **Funding**: Natalia Díaz-Rodríguez is supported by grant IJC2019-039152-I funded by MCIN/AEI /10.13039/501100011033, by "ESF Investing in your future", Google Research Scholar Program, and by the European Union through the Marie Curie Postdoctoral Fellowship (Project: 101059332 — RRR-XAI — HORIZON-MSCA-2021-PF-01). J. Del Ser is supported by the Basque Government through the ELKARTEK program and the consolidated research group MATHMODE (IT1456-22). We thank D. H. S. Maithripala and Elisa Massi for valuable feedback on an early version of this article.

- **Conflict of Interest**: J. Del Ser is an editorial board member of Cognitive Computation. The authors declare that they have no other conflict of interest regarding this work. Views and opinions expressed are however those of the author(s) only and do not necessarily reflect those of the European Union or University of Granada. Neither the European Union nor the granting authority can be held responsible for them.

- **Ethical approval**: This article does not contain any studies with human participants or animals performed by any of the authors.

**Data Availability Statement**

All the experiments were exclusively conducted on open source software, namely pytorch [55], roboschool [53], OpenAI gym [52]. The implementation, requirements and configurations to run the experiments have been provided in the following reproducibility checklist.

- **Implementation**:
  https://github.com/punk95/Continual-Learning-With-Curiosity

- **Hyper Parameter Configuration**: Hyper-Parameter values were specified in Appendix B

- **Hardware requirements**: All the algorithms were run on a 12 core AMD Ryzen™ 9 5900X processor and an Nvidia RTX 3080 GPU.

- **Number of Algorithms run**: Each of the results were obtained by averaging over 8 experiments with different seeds.

- **Seeds**: A different randomly generated seed was used for every individual experiment.

- **Statistics used for the results**: We have used the discounted reward and the composition ratio of the buffers as our statistic to evaluate the algorithms. The results were averaged over 8 runs.

# Appendices

## A. Implementation

Pytorch [55] based implementation of all the implementations (HCB (Section 2.1), HRBTS (Section 2.2), HRB [38], MTR [39] and FIFO) and the experiments used in work can be accessed at https://github.com/punk95/Continual-Learning-With-Curiosity

## B. Hyper Parameters of the Curiosity Based Algorithms and Experiments

Continual learning setup itself was emulated in the environment by changing certain parameters in the control task. In both the Walker2D and the Hopper environments power parameter of the learner as defined in [53] was changed, whereas in the Pendulum environment the length of the Pendulum was changed. In the Pendulum's case length was changed within the range of 1.0 to 1.8 in-environment units. As per Hopper's case power ranged from 0.75 to 8.75 in-environment units. On the other hand, Walker2D's power ranged from 1.4 to 13.4 in-environment units. Here the learning agents were not provided with explicit knowledge of the change. Except for the every timestep task encounter with small task drift and task re-visitation setting, tasks were defined by the minimum, maximum, and mean of the range. For example, three tasks in the Pendulum's case correspond to learning to balance a Pendulum corresponding to the lengths of 1.0, 1.4, and 1.8 in-environment units.

We believe that certain characteristics should be maintained when selecting these parameter ranges (for the prolonged task encounter with substantial task drift, and no task re-visitation and frequent, short-lived task encounter with substantial task drift and task re-visitation settings) in order to create meaningful experiments, which are in line with the scope of this paper. **1.** *Since neural architectural changes are outside the scope of this paper, the range should be selected such that a single architecture, to a certain extent, can generalize well across these tasks.* **2.** *Tasks selected should be independent of each other (at least to some level).* This aspect is essential as one can misinterpret generalization by predominantly training on a single task when that task is dominant in time. **3.** *The tasks should be immune to few-shot learning when they are trained with a network that was optimized for a previous task.* To elaborate further when different tasks are close to each other by nature, if we are to train on subsequent tasks starting with a network that is optimized for the previous task sometimes we may only need to see a few transition tuples of the new task for the current network to start performing well (to start adjusting for the new task). This is indeed possible only by the fact that these tasks by nature share some kind of similarity between each other and we can't expect this phenomenon to appear in general. If we are to choose two different tasks which are similar/easily adaptable as mentioned before then we may misinterpret the results, which are a consequence of the algorithm predominately training on one task, as the algorithm performing well in generalization when in fact it is an one off scenario and the algorithm may perform worse on other scenarios. Thus it becomes essential to choose tasks such that they are not very similar in nature. To these ends, via trial and error, we selected these aforementioned ranges so that the tasks defined by the minimum, maximum, and mean of these ranges would suffice these characteristics.

*B.1. Parameter tuning:*

| Environment | Pendulum | Hopper | Walker2D |
|---|---|---|---|
| *1. Scenario Specific Parameters* | | | |
| Max steps per episode | 200 | 1000 | 1000 |
| Environment variable changed | length | power | power |
| Variable changed at (time step) | [0, 20000, 120000] | [0, 50000, 350000] | [0, 250000, 350000] |
| Changed variable value | [1.0, 1.4, 1.8] | [0.75, 4.75, 8.75] | [1.4, 7.4, 13.4] |
| *2. General Algorithm Specific Parameters* | | | |
| Base Algorithm | SAC [36] | SAC [36] | SAC [36] |
| Batch size | 512 | 512 | 512 |
| FIFO fraction | 0.05 | 0.05 | 0.05 |
| Hidden Layer | [256, 256] | [256, 256] | [256, 256] |
| Total no steps | 150000 | 400000 | 400000 |
| Total Buffer size | 20000 | 50000 | 100000 |
| No of Curiosity Networks | 3 | 3 | 1 |
| Forward, Inverse, Reward Curiosity weights | [0,1,0] | [0,1,0] | [0,1,0.05] |
| Reward Curiosity weights | | | |
| *3. HRBTS (Section 2.2) Buffer Specific Parameters* | | | |
| $n$, $k$, $m$ | 600, 8000, 1.5 | 500, 30000, 2.5 | 2000, 30000, 0.2 |

Table 2: Hyper Parameter for all the Algorithms (HCB (Section 2.1), HRBTS (Section 2.2), HRB [38], MTR [39] and FIFO) in Pendulum, Hopper and Walker2D environments.

*B.2. Parameters: Every timestep task encounter with small task drift and task re-visitation*

| Environment | Pendulum | Hopper |
|---|---:|---:|
| *1. Scenario Specific Parameters* | | |
| Max steps ($t$) per episode | 200 | 1000 |
| Environment variable changed | length | power |
| Formula for Variable change | $l_{min} + \sin(t.10^{-4}).\delta_l$ | $p_{min} + \sin(t.10^{-4}).\delta_p$ |
| $\delta_l, \delta_p$ values | $\delta_l = (l_{max} - l_{min})$ | $\delta_p = (p_{max} - p_{min})$ |
| Max ($l_{max}, p_{max}$) and Min ($l_{min}$ or $p_{min}$) Values | 1.0 and 1.8 | 0.75, 8.75 |
| *2. General Algorithm Specific Parameters* | | |
| Base Algorithm | SAC [36] | SAC [36] |
| Batch size | 512 | 512 |
| FIFO fraction | 0.05 | 0.05 |
| Hidden Layer | [256, 256] | [256, 256] |
| Total no steps | 150000 | 400000 |
| Total Buffer size | 20000 | 50000 |
| No of Curiosity Networks | 1 | 1 |
| Forward, Inverse, Reward Curiosity weights | [0,1,0] | [0,1,0] |
| *3. HRBTS Section 2.2 Buffer Specific Parameters* | | |
| $n, l, m$ | 600, 8000, 1.5 | 500, 30000, 2.5 |

Table 3: Hyper Parameter for all the Algorithms (HCB (Section 2.1), HRBTS (Section 2.2), HRB [38], MTR [39] and FIFO) in Pendulum and Hopper environments.

*B.3. Parameters: Frequent, short-lived task encounter with substantial task drift and task re-visitation*

| Environment | Pendulum | Hopper |
|---|---:|---:|
| Max steps per episode | 200 | 1000 |
| Environment variable changed | length | power |
| Variable changed at (time step) | [0, 32275, 37275, 56602, 61602, 81694, 86694, 105977, 110977, 118155, 123155, 141158, 141158, 146158]. | [0, 86067, 98567, 150939, 163439, 217851, 230351, 282606, 295106, 315080, 327580, 376422, 388922,] |
| Changed variable value | [1.4, 1.0, 1.4, 1.8, 1.4, 1.0, 1.4, 1.0, 1.4, 1.8, 1.4, 1.8 1.4] | [4.75, 0.75, 4.75, 8.75, 4.75, 0.75, 4.75, 0.75, 4.75, 8.75, 4.75, 8.75, 4.75] |
| **2. General Algorithm Specific Parameters** | | |
| Base Algorithm | SAC [36] | SAC [36] |
| Batch size | 512 | 512 |
| FIFO fraction | 0.05 | 0.05 |
| Hidden Layer | [256, 256] | [256, 256] |
| Total no steps | 150000 | 400000 |
| Total Buffer size | 20000 | 50000 |
| No of Curiosity Networks | 1 | 1 |
| Forward, Inverse, Reward Curiosity weights | [0,1,0] | [0,1,0] |
| **3. HRBTS (Section 2.2) Buffer Specific Parameters** | | |
| $n, l, m$ | 600, 8000, 1.5 | 500, 30000, 2.5 |

Table 4: Hyper Parameter for all the Algorithms (HCB (Section 2.1), HRBTS (Section 2.2), HRB [38], MTR [39] and FIFO) in Pendulum and Hopper environments.

## C. Additional Results and Analyses

*C.1. Prolonged task encounter with substantial task drift, and no task re-visitation*

| Phase 1 (Task 1) | FIFO | HRB | MTR | HRBTS (ours) | HCB (ours) |
|---|---|---|---|---|---|
| Task 1's Reward | -230.7 ± 49.5 | -262.3 ± 69.6 | -235.5 ± 45.2 | -256.5 ± 39.1 | **-229.2 ± 50.2** |
| Task 2's Reward (dominant) | -839.9 ± 133.9 | -799.3 ± 70.2 | -819.1 ± 120.5 | -838.5 ± 68.7 | -842.8 ± 110.3 |
| Task 3's Reward | -941.7 ± 79.2 | -934.0 ± 75.0 | -914.5 ± 99.2 | -938.3 ± 81.0 | -910.0 ± 94.1 |
| Phase 2 (Task 2) | FIFO | HRB | MTR | HRBTS (ours) | HCB (ours) |
| Task 1's Reward | -911.9 ± 90.1 | -835.5 ± 87.6 | -791.9 ± 64.9 | **-368.8 ± 205.0** | -795.9 ± 168.8 |
| Task 2's Reward (dominant) | **-370.2 ± 46.8** | -385.0 ± 90.1 | -371.7 ± 55.1 | -404.9 ± 110.7 | -382.4 ± 74.8 |
| Task 3's Reward | -767.9 ± 46.1 | -795.1 ± 81.4 | -779.7 ± 74.4 | -816.5 ± 74.3 | -830.7 ± 50.2 |
| Phase 3 (Task 3) | FIFO | HRB | MTR | HRBTS (ours) | HCB (ours) |
| Task 1's Reward | -1041.3 ± 70.4 | -953.7 ± 86.8 | -1005.0 ± 72.8 | -999.7 ± 61.5 | **-995.4 ± 64.7** |
| Task 2's Reward (dominant) | -782.2 ± 153.7 | -399.3 ± 98.2 | -539.1 ± 197.6 | -408.3 ± 96.9 | **-374.9 ± 76.0** |
| Task 3's Reward | **-511.5 ± 113.4** | -748.7 ± 81.4 | -653.2 ± 124.8 | -730.1 ± 110.3 | -729.9 ± 66.8 |

Table 5: Rewards for Pendulum in all phases. Here Task 1 corresponds to Pendulum's length of 1.0, Task 2 corresponds to Pendulum's length of 1.4 and Task 3 corresponds to Pendulum's length of 1.8. Phase 1 corresponds to t=0 to t = 30000 where the agent was exposed to Task 1. Phase 2 corresponds to t = 30000 to t = 130000 where the agent was exposed to Task 2. Phase 3 corresponds to t = 130000 to t = 150000 where the agent was exposed to Task 3. Here the dominant task corresponds to the task that is the most exposed to the agent (that task at which the agent spent most of it's time training on) during the training time

| Phase 1 (Task 1) | FIFO | HRB | MTR | HRBTS (ours) | HCB (ours) |
|---|---|---|---|---|---|
| Task 1's Reward | 1268.2 ± 531.9 | 901.7 ± 561.7 | **1537.0 ± 149.9** | 958.7 ± 423.7 | 1447.9 ± 509.2 |
| Task 2's Reward (dominant) | -19.9 ± 22.5 | -8.3 ± 13.2 | -12.6 ± 20.0 | -17.3 ± 25.2 | -16.0 ± 25.7 |
| Task 3's Reward | -6.7 ± 22.9 | -13.7 ± 23.4 | -20.0 ± 23.6 | -19.2 ± 30.3 | -20.8 ± 28.4 |
| Phase 2 (Task 2) | FIFO | HRB | MTR | HRBTS (ours) | HCB (ours) |
| Task 1's Reward | 274.7 ± 166.2 | 374.3 ± 329.4 | 233.5 ± 264.1 | **464.1 ± 323.3** | 318.7 ± 278.2 |
| Task 2's Reward (dominant) | 1235.4 ± 733.3 | **1419.2 ± 719.7** | 1084.9 ± 615.2 | 799.9 ± 529.1 | 438.9 ± 326.1 |
| Task 3's Reward | -15.3 ± 27.9 | 41.3 ± 40.2 | 72.7 ± 55.9 | 32.6 ± 52.1 | 53.1 ± 39.6 |
| Phase 3 (Task 3) | FIFO | HRB | MTR | HRBTS (ours) | HCB (ours) |
| Task 1's Reward | 298.9 ± 297.2 | 188.6 ± 115.5 | 59.5 ± 49.2 | **491.3 ± 376.5** | 312.2 ± 209.4 |
| Task 2's Reward (dominant) | 472.3 ± 303.7 | **1371.7 ± 753.7** | 1347.6 ± 835.2 | 563.1 ± 554.8 | 355.6 ± 252.3 |
| Task 3's Reward | 586.0 ± 366.5 | 336.4 ± 202.6 | **1015.5 ± 576.5** | 326.2 ± 266.7 | 147.7 ± 97.6 |

Table 6: Rewards for Hopper in all phases. Phase 1 corresponds to t=0 to t = 50000 where the agent was exposed to Task 1. Here Task 1 corresponds to Hopper's power of 0.75, Task 2 corresponds to Hopper's power of 4.75 and Task 3 corresponds to Hopper's power of 8.75. Phase 2 corresponds to t = 50000 to t = 350000 where the agent was exposed to Task 2. Phase 3 corresponds to t = 350000 to t = 400000 where the agent was exposed to Task 3. Here the dominant task corresponds to the task that is the most exposed to the agent (that task at which the agent spent most of it's time training on) during the training time

| Phase 1 (Task 1) | FIFO | HRB | MTR | HRBTS (ours) | **HCB (ours)** |
| --- | --- | --- | --- | --- | --- |
| Task 1's Reward *(dominant)* | 1641.0 ± 363.4 | **1825.9 ± 295.0** | 1279.6 ± 866.1 | 1666.2 ± 382.6 | 1585.3 ± 570.5 |
| Task 2's Reward | -42.2 ± 19.0 | -53.2 ± 16.7 | 187.9 ± 428.7 | -57.4 ± 12.9 | -34.3 ± 22.0 |
| Task 3's Reward | -39.1 ± 31.8 | -22.7 ± 19.7 | 11.7 ± 64.9 | -48.6 ± 24.9 | -27.2 ± 18.3 |
| *Phase 2 (Task 2)* | *FIFO* | *HRB* | *MTR* | *HRBTS (ours)* | *HCB (ours)* |
| Task 1's Reward | 207.3 ± 167.1 | 444.0 ± 150.2 | 374.6 ± 148.5 | **445.8 ± 425.2** | 185.9 ± 167.6 |
| Task 2's Reward *(dominant)* | **1558.7 ± 402.8** | 380.8 ± 124.5 | 1500.2 ± 569.9 | 845.0 ± 167.4 | 276.3 ± 193.6 |
| Task 3's Reward | -67.3 ± 36.2 | -48.2 ± 21.5 | -30.3 ± 53.1 | -77.8 ± 22.7 | -13.2 ± 28.3 |
| *Phase 3 (Task 3)* | *FIFO* | *HRB* | *MTR* | *HRBTS (ours)* | *HCB (ours)* |
| Task 1's Reward | 137.2 ± 147.6 | **307.6 ± 182.3** | 127.1 ± 147.4 | 95.6 ± 138.0 | 168.1 ± 172.5 |
| Task 2's Reward | **2145.9 ± 259.5** | 52.5 ± 27.2 | 172.5 ± 238.4 | 64.1 ± 74.8 | 55.7 ± 30.8 |
| Task 3's Reward *(dominant)* | **740.2 ± 427.3** | 5.4 ± 8.3 | 394.0 ± 268.6 | 152.8 ± 241.3 | 489.6 ± 222.6 |

Table 7: Rewards for Walker 2D in all phases. Phase 1 corresponds to t=0 to t = 250000 where the agent was exposed to Task 1. Here Task 1 corresponds to Walker's power of 1.40, Task 2 corresponds to Walker's power of 7.40 and Task 3 corresponds to Walker's power of 13.40. Phase 2 corresponds to t = 250000 to t = 350000 where the agent was exposed to Task 2. Phase 3 corresponds to t = 350000 to t = 400000 where the agent was exposed to Task 3. Here the dominant task corresponds to the task that is the most exposed to the agent (that task at which the agent spent most of it's time training on) during the training time

*C.2. Frequent, short-lived task encounter with substantial task drift and task re-visitation*

| Phase 1 (Task 3) | FIFO | HRB | MTR | HRBTS (ours) | HCB (ours) |
| --- | --- | --- | --- | --- | --- |
| Task 1's Reward | -460.4 ± 245.3 | -539.3 ± 200.7 | -557.4 ± 233.5 | **-354.4 ± 151.6** | -425.2 ± 200.7 |
| Task 2's Reward *(dominant)* | -410.0 ± 70.7 | **-325.4 ± 57.5** | -378.2 ± 93.6 | -350.0 ± 97.5 | -460.7 ± 57.5 |
| Task 3's Reward | -776.7 ± 96.0 | -768.9 ± 67.5 | **-738.1 ± 94.4** | -753.4 ± 80.3 | -786.4 ± 67.5 |
| *Phase 2 (Task 1)* | *FIFO* | *HRB* | *MTR* | *HRBTS (ours)* | *HCB (ours)* |
| Task 1's Reward | -685.3 ± 191.5 | -520.5 ± 196.6 | -755.2 ± 198.6 | -492.9 ± 243.9 | **-436.2 ± 196.6** |
| Task 2's Reward *(dominant)* | -430.6 ± 61.5 | -347.7 ± 55.1 | **-342.8 ± 65.4** | -395.9 ± 85.1 | -429.0 ± 55.1 |
| Task 3's Reward | -770.0 ± 76.1 | -770.8 ± 102.5 | -804.5 ± 91.3 | -806.0 ± 80.5 | **-743.1 ± 102.5** |
| *Phase 3 (Task 2)* | *FIFO* | *HRB* | *MTR* | *HRBTS (ours)* | *HCB (ours)* |
| Task 1's Reward | -879.6 ± 195.0 | -682.9 ± 166.9 | -935.9 ± 56.5 | -809.0 ± 145.7 | **-552.8 ± 166.9** |
| Task 2's Reward *(dominant)* | **-340.5 ± 109.4** | -383.9 ± 81.6 | -346.0 ± 40.0 | -346.8 ± 66.1 | -412.4 ± 81.6 |
| Task 3's Reward | -762.3 ± 76.0 | -764.1 ± 56.4 | **-731.4 ± 121.0** | -840.6 ± 70.5 | -784.3 ± 56.4 |

Table 8: Rewards for Pendulum in all phases. Here Task 1 corresponds to Pendulum's length of 1.0, Task 2 corresponds to Pendulum's length of 1.4 and Task 3 corresponds to Pendulum's length of 1.8. Phase 1 corresponds to t=56602 to t = 61602 where the agent was exposed to Task 3. Phase 2 corresponds to t = 105977 to t = 110977 where the agent was exposed to Task 1. Phase 3 corresponds to t = 123155 to t = 141158 where the agent was exposed to Task 2. Here the dominant task corresponds to the task that is the most exposed to the agent (that task at which the agent spent most of it's time training on) during the training time

| Phase 1 (Task 1) | FIFO | HRB | MTR | HRBTS (ours) | HCB (ours) |
|---|---|---|---|---|---|
| Task 1's Reward | 44.3 ± 16.1 | 60.7 ± 39.8 | 55.2 ± 24.5 | 50.1 ± 37.4 | **78.2 ± 39.8** |
| Task 2's Reward (dominant) | **220.9 ± 62.5** | 198.9 ± 46.2 | 196.6 ± 183.1 | 206.3 ± 109.8 | 173.1 ± 46.2 |
| Task 3's Reward | **377.4 ± 206.6** | 260.2 ± 108.7 | 268.8 ± 146.1 | 254.3 ± 121.7 | 241.2 ± 108.7 |
| *Phase 2 (Task 2)* | FIFO | HRB | MTR | HRBTS (ours) | HCB (ours) |
| Task 1's Reward | 42.7 ± 19.1 | 71.1 ± 36.4 | 65.8 ± 32.3 | 96.7 ± 115.9 | **105.5 ± 36.4** |
| Task 2's Reward (dominant) | **508.5 ± 287.0** | 239.2 ± 61.6 | 297.8 ± 197.2 | 243.8 ± 100.4 | 261.8 ± 61.6 |
| Task 3's Reward | **445.3 ± 193.3** | 288.1 ± 135.4 | 190.3 ± 128.0 | 218.5 ± 128.2 | 201.5 ± 135.4 |
| *Phase 3 (Task 3)* | FIFO | **HRB** | MTR | HRBTS (ours) | HCB (ours) |
| Task 1's Reward | 122.2 ± 63.3 | 180.5 ± 146.0 | **276.9 ± 226.7** | 98.2 ± 75.8 | 80.9 ± 146.0 |
| Task 2's Reward (dominant) | **-340.5 ± 109.4** | -383.9 ± 81.6 | -346.0 ± 40.0 | -346.8 ± 66.1 | -412.4 ± 81.6 |
| Task 3's Reward | 12.0 ± 27.2 | **219.0 ± 116.2** | 74.1 ± 45.3 | 128.2 ± 75.9 | 121.3 ± 116.2 |

Table 9: Rewards for Hopper in all phases. Here Task 1 corresponds to Hopper's power of 0.75, Task 2 corresponds to Hopper's power of 4.75 and Task 3 corresponds to Hopper's power of 8.75. Phase 1 corresponds to t=86067 to t = 111067 where the agent was exposed to Task 1. Phase 2 corresponds to t = 111067 to t = 150939 where the agent was exposed to Task 2. Phase 3 corresponds to t = 315080 to t = 327580 where the agent was exposed to Task 3. Here the dominant task corresponds to the task that is the most exposed to the agent (that task at which the agent spent most of its time training on) during the training time